\documentclass[11pt,a4paper]{article}
\usepackage[pdfborder={0 0 0}]{hyperref}
\usepackage{cmap}
\usepackage[utf8x]{inputenc}
\usepackage[T1]{fontenc}
\usepackage[english]{babel}
\usepackage{amssymb,amsmath}
\usepackage{amsthm}
\usepackage{bbm}
\usepackage{microtype}
\usepackage{lmodern}

{
\rmfamily
\DeclareFontShape{T1}{lmr}{m}{sc}{<->ssub*cmr/m/sc}{}
\DeclareFontShape{T1}{lmr}{b}{sc}{<->ssub*cmr/b/sc}{}
\DeclareFontShape{T1}{lmr}{bx}{sc}{<->ssub*cmr/bx/sc}{}
}

\newcommand{\thmheadercommand}[1]{\textbf{\scshape{}#1.\\*}}

\newtheoremstyle{yannthm}{\topsep}{\topsep}{\slshape}{}{\scshape\bfseries}{.}{.5em}{%
\thmname{#1}\thmnumber{ #2}\thmnote{#3}%
}
\newtheoremstyle{yannthm2}{\topsep}{\topsep}{}{}{\scshape\bfseries}{.}{.5em}{%
\thmname{#1}\thmnumber{ #2}\thmnote{#3}%
}

\def\d{\operatorname{d}\!{}}

\def\R{{\mathbb{R}}}

\renewcommand{\geq}{\geqslant}
\renewcommand{\leq}{\leqslant}

\newcommand{\deq}{\mathrel{\mathop:}=}
\newcommand{\eqd}{=\mathrel{\mathop:}}

\def\eps{\varepsilon}
\renewcommand{\epsilon}{\varepsilon}
\renewcommand{\phi}{\varphi}

\DeclareMathOperator{\Var}{Var}

\let\oldPr\Pr
\renewcommand{\Pr}{\oldPr\nolimits}
\newcommand{\E}{\mathbb{E}}
\newcommand{\KL}[2]{\mathrm{KL}\!\left(#1 \,|\hspace{-.15ex}|\,#2\right)}

\newcommand{\abs}[1]{\left|\mskip1mu#1\right|}
\newcommand{\norm}[1]{\left\|#1\right\|}

\newcommand{\1}{\mathbbm{1}}

\DeclareMathOperator*{\argmax}{arg\,max}

{
\theoremstyle{yannthm}
\newtheorem{defi}{Definition}
\newtheorem*{defi*}{Definition}
\newtheorem{prop}[defi]{Proposition}
\newtheorem*{prop*}{Proposition}

\newtheorem*{thm*}{Theorem}

\newtheorem*{lem*}{Lemma}

\newtheorem*{cor*}{Corollary}

\newtheorem*{ex*}{Example}

\newtheorem*{subenonce}{}

\theoremstyle{yannthm2}

\newtheorem*{exo*}{Exercise}

\newtheorem{rem}[defi]{Remark}
\newtheorem*{rem*}{Remark}

\newtheorem*{subenonce2}{}

}

\newcommand{\transp}[1]{#1^{\!\top}\!}

\usepackage{url}
\usepackage{float}
\usepackage{graphicx}
\usepackage{stmaryrd}
\usepackage{multirow}

\newcommand{\e}{\mathrm{e}}
\newcommand{\network}{\mathcal{N}}
\newcommand{\A}{\mathcal{A}}
\newcommand{\actf}{s}
\newcommand{\natnorm}[1]{\norm{#1}_{\mathrm{nat}}}
\newcommand{\bpnorm}[1]{\norm{#1}_{\mathrm{bp}}}
\newcommand{\rbpnorm}[1]{\norm{#1}_{\mathrm{rbp}}}
\newcommand{\uopnorm}[1]{\norm{#1}_{\mathrm{uop}}}
\newcommand{\ruopnorm}[1]{\norm{#1}_{\mathrm{ruop}}}
\newcommand{\ffnorm}[1]{\norm{#1}_{\mathrm{ff}}}
\newcommand{\rffnorm}[1]{\norm{#1}_{\mathrm{rff}}}
\newcommand{\rnatnorm}[1]{\norm{#1}_{\mathrm{rnat}}}
\newcommand{\tsum}{{\textstyle \sum}}
\newcommand{\D}{\mathcal{D}}

\renewcommand{\d}{\ensuremath{\hspace{0.05em}\delta\hspace{-.05em}}}

\newcommand{\intint}[1]{\llbracket #1 \llbracket}

\newfloat{Example}{h}{ex}

\title{Riemannian metrics for neural networks II: recurrent networks and
learning symbolic data sequences}
\author{Yann Ollivier}

\begin{document}

\maketitle

\begin{abstract}
Recurrent neural networks are powerful models for sequential data, able
to represent complex dependencies in the sequence that simpler models such as hidden
Markov models cannot handle. Yet they are notoriously hard to train.

Here we introduce a training procedure using a gradient ascent in a
Riemannian metric:
this produces an
algorithm independent from design choices such as the
encoding of parameters and unit activities. This metric gradient ascent is
designed to have an algorithmic cost close to backpropagation through
time for sparsely connected networks.

We use this procedure on \emph{gated leaky neural networks} (GLNNs),
a variant of recurrent neural networks with an architecture
inspired by finite automata and an evolution equation inspired by
continuous-time networks.

GLNNs trained with a Riemannian gradient are demonstrated to effectively capture a variety of
structures in synthetic problems: basic block nesting
as in context-free grammars (an important feature of natural languages,
but difficult to learn), intersections of multiple independent
Markov-type relations, or long-distance relationships such as the
distant-XOR problem.

This method does not require adjusting the network structure or initial
parameters:
the network used is a sparse random graph and the
initialization is identical for all problems considered.
\end{abstract}


The problem considered here is to learn a probabilistic model for an
observed sequence of symbols $(x_0,\ldots,x_t,\ldots)$ over a finite
alphabet $\A$. Such a model can be used for prediction,
compression, or generalization.

Hidden Markov models (HMMs) are frequently used in such a setting. However, the
kind of algorithmic structures HMMs can represent is limited because of
the underlying finite automaton structure. 
Examples of simple sequential data that cannot be, or cannot conveniently
be, represented by HMMs are discussed below; for instance, subsequence insertions,
or intersections of multiple independent
constraints.

Recurrent neural networks (RNNs) are an alternative with higher modelling
power. However, their training comes with its own limitations; in
particular, picking long-distance dependencies remains problematic
\cite{Bengio_longdistance94,HochreiterSchmidhuber1997,Jaeger_tutorial}.
Techniques to deal with this problem include long short-term memory
(LSTM) networks \cite{HochreiterSchmidhuber1997} or echo state networks
(ESN) \cite{Jaeger_tutorial}.

Here we use a new training procedure
which realizes a gradient ascent using a suitable Riemannian
\emph{metric},
instead of backpropagation, at a small computational cost. Moreover, we
use
\emph{gated leaky neural networks} (GLNNs),
a variation on the RNN architecture.
More precisely:

\begin{itemize}
\item Rather than standard backpropagation through time, for training
the model we use a
gradient inspired by Riemannian geometry, using \emph{metrics} for neural
networks 
as introduced in \cite{gradnn}, adapted to a recurrent context. This
makes learning less sensitive to arbitrary design choices, and provides
a substantial improvement in learning speed and quality. An important point is
doing so while keeping a scalable algorithm. Here the asymptotic algorithmic complexity is
identical to backpropagation through time for sparsely connected
networks.

\item In GLNNs, at each time in the production of a sequence of symbols, the neural
network weights depend on the symbol last produced (``gated'' units). This
is inspired by finite automata in which the next state depends both on
the current state and the currently produced symbol, and allows for an easy
representation of automaton-like structures. Such ``gated'' models
have already been used, e.g., in \cite{Sutskever_text2011}, and arguably the
LSTM architecture.

\item The dynamics of GLNNs is modified in a way inspired by
continuous-time (or ``leaky'') neural networks: the connection weights between the units control
the \emph{variation} of the activation levels, rather than directly
setting the activation levels at the next step. This provides an
integrating effect and is efficient, for instance, at modelling some hierarchical,
context-free-grammar--like structures in which an internal state must be
held constant while something else is happening.
\end{itemize}

Much of this text is devoted to the derivation of
Riemannian
metrics for recurrent networks. Indeed, we believe the use of a proper gradient is
a major ingredient for an effective learning procedure. The standard gradient
ascent update over a parameter $\theta$ can be seen as a way to increase the value of a function
$f(\theta)$ while changing as least as possible the numerical value
$\theta$:
\begin{equation}
\theta'=\theta+\eta \frac{\partial f}{\partial \theta}
\qquad \Rightarrow \qquad
\theta'\approx \argmax_{\theta'}
\left\{f(\theta')-\frac{1}{2\eta}\norm{\theta-\theta'}^2\right\}
\end{equation}
for small enough learning rates $\eta$ (where $\approx$ means ``up to
$O(\eta^2)$ when $\eta\to 0$''). The norm $\norm{\theta-\theta'}$
depends on how the parameters are cast as a set of real numbers.
If,
instead, one uses a measure of distance between $\theta$ and $\theta'$
depending on what the network \emph{does}, rather than how the numbers in
$\theta$ and $\theta'$ differ, the penalty for moving $\theta$ in
different directions becomes different and hopefully yields better
learning. One possible benefit, for instance, is self-adaptation 
of the cost of moving
$\theta$ in certain directions, depending on the current behavior of the
network.
Another benefit is \emph{invariance} of the learning procedure from a
number of designing choices, such as using a logistic or tanh activation
function, or scaling the values of parameters (choices which affect the
conventional gradient ascent).

The primary example of an invariant gradient ascent is Amari's \emph{natural gradient}, which amounts to
replacing $\norm{\theta-\theta'}^2$ with the Kullback--Leibler divergence
$\KL{\Pr_\theta}{\Pr_{\theta'}}$
between the distributions defined by the network (seen as a probabilistic
model of the data).  However, the
natural gradient comes at a great algorithmic cost.  ``Hessian-free''
techniques \cite{Martens2010,Martens2011,Martens2012} allow to
approximate it to some extent and have yielded good results, but are
still quite computationally expensive.

Here we build two metrics for recurrent
neural networks having some of the key properties of the natural gradient, but
at a computational cost closer to that of backpropagation through time.
The resulting algorithm is first presented in Section~\ref{sec:algo} in
its final form. The algorithm might look arbitrary at first sight, but is
theoretically well-grounded;
in Sections~\ref{sec:metricsintro}--\ref{sec:bpm} we derive it
step by step from the principles in \cite{gradnn} adapted to a recurrent
setting.

This construction builds on the Riemannian geometry framework for neural
networks from \cite{gradnn}. The activities of units in the network
are assumed to belong to a \emph{manifold}: intuitively, they represent
``abstract quantities'' representable by numbers, but no preferred
correspondence with $\R$ is fixed. This forces us to write only
\emph{invariant} algorithms which do not depend on the chosen numerical
representation of the activities. Such algorithms are more impervious to
design choices (e.g., changing the activation function from logistic to
tanh has no effect); as a consequence, if they work well on one problem,
they will tend to work well on rewritings of the same problem using
different numerical representations. Thus, such algorithms are more
``agnostic'' as to physical meaning of the activities of the units
(activation levels, activation frequencies, log-frequencies, ...).

\begin{rem}
The three changes introduced above with respect to standard RNNs are
independent and can be used
separately. For instance, the metrics can be used for
any network architecture.
\end{rem}

\begin{rem}
The approach is not specific to symbolic sequences: instead of transition
parameters $\tau_{ijx_t}$ depending on the latest symbol $x_t$, one can
use transition weights which depend on the components of the latest
input vector $x_t$.
\end{rem}

\begin{rem}
The gradient update proposed is independent of the training example management
scheme (batch, online, small batches, stochastic gradient\ldots).
\end{rem}

\begin{rem}
The algorithm presented here is quadratic in network connectivity (number
of connections per unit), and we have used it with very sparse networks
(as few as 3 connections per unit), which apparently perform well. For non-sparse
networks, a version with complexity
linear in the number of connections, but with fewer invariance
properties, is presented at the end of
Section~\ref{sec:algo}.
\end{rem}

\paragraph{Examples.}
Let us present a few examples of data that we have found
can be efficiently learned by GLNNs. 
Other techniques that have been used to deal with such sequences include
long short-term memory (LSTM) networks \cite{HochreiterSchmidhuber1997}
(see for instance \cite{Graves2013} for a recent application using
stacked LSTMs for text modelling) and
echo state networks (ESN) \cite{Jaeger_tutorial}.
Here we do not have to engineer a particular network structure or to have
prior knowledge of the scale of time correlations for initialization:
in our experiments the network is a sparse random graph and parameter
initialization is the same for all problems.

\newcommand{\dataexstyle}[1]{{\noindent\small{\texttt{#1}}}}

Example~\ref{ex:insert} illustrates a type of operation frequent in
natural languages (and artificial programming languages): in the course of
a sequence, a subsequence is inserted, then the main sequence resumes
back exactly where it was interrupted. This kind of structure is impossible to
represent within a Markovian model, and is usually modelled with context-free
grammars (the learning of which is still problematic).

In this example, the main sequence is the Latin alphabet. Sometimes a
subsequence is inserted which spells out the digits from 0 to 9. In this
subsequence, sometimes a subsubsequence is inserted containing nine
random (to prevent rote learning) capital letters
(Example~\ref{ex:insert}).

\begin{Example}
\dataexstyle{abcdefghijklmnopqrs(01[HSATXUEUZ]2[OYNFIWWOR]345[ZYMBOMYBZ]6789)tuvwxyz
\\
abcde(01234567[FFRLCMKVI]89)fghijklmnopqrstuvwxyz
\\...}
\caption{Inserting subsequences, a simple context-free grammar.}
\label{ex:insert}
\end{Example}

Here the difficulty, both for HMMs and recurrent neural
networks trained by ordinary backpropagation through time, is in starting again at the right point after the interruption
caused by the subsequence.

\bigskip

Example~\ref{ex:xor} is a pathological synthetic problem traditionally
considered among the hardest for recurrent neural networks (although it
can be represented by a simple finite automaton): the distant XOR
problem. In a random binary sequence, two positions are marked at random
(here with the symbol \texttt{X}), and the binary
symbol at the end of each line is the logical XOR of the two random bits
following the \texttt{X} marks.
Use of the XOR function prevents detecting a correlation between
the XOR result and any one of the two arguments.

\begin{Example}
\dataexstyle{ 1 1 1 0 0X1 1X1 1 1 1 1 0 0 1 0 1 0 0 0 1 0 1 0 0 1=0
\\\phantom{X}0X1X0 1 1 0 1 0 1 1 1 0 1 0 1 0 0 1 0 0 1 0 0 1 0 0=1
\\...
}
\caption{Long-distance XOR.}
\label{ex:xor}
\end{Example}

On this example, apparently the best performance for RNNs is obtained in
\cite{Martens2011}: with 100 random bits on each line, the failure rate
is about $75\%$, where ``failure'' means that a run examines more than 50
million examples before reaching an error rate below $1\%$ \cite[legend
of Figure~3]{Martens2011}.

\bigskip

Example~\ref{ex:mus} is synthetic music notation (here in LilyPond
format\footnote{\url{http://lilypond.org/}}), meant to illustrate the
intersection of several independent constraints. Successive musical bars
are separated by a \texttt{|} symbol. Each bar is a succession of notes
separated by spaces, where each note is made of a pitch (\texttt{a},
\texttt{b}, \texttt{c}, ...) and value (\texttt{4} for a quarter note,
\texttt{2} for a half note, \texttt{4.}\ for a dotted quarter note,
etc.). In each bar, a hidden variable with three possible values
determines a \emph{harmony} which restricts the possible pitches used in
this bar. Harmonies in successive bars follow a specific deterministic
pattern. Additionally, in each bar, the successive durations are taken
from a finite set of possibilities (rhythms commonly encountered in
waltzes). Rhythm is chosen independently from pitch and harmony.
The resulting probability distribution is the intersection of all these
constraints.

\begin{Example}
\dataexstyle{c2 c4 |}
\dataexstyle{f4.\ a8 c4 |}
\dataexstyle{g4 b4 g8 d8 |}
\dataexstyle{g4.\ g8 g4 |}
\dataexstyle{e4 c4 c4 | ...}
\caption{Synthetic music.}
\label{ex:mus}
\end{Example}

This example can be represented as a Markov chain, but only using a
huge state space. The ``correct'' representation of the constraints is
more compact, which allows for efficient learning, whereas a Markov
representation would essentially need to see every possible combination
of rhythm and pitches to learn the underlying structure.

\bigskip

Example~\ref{ex:anbn} is the textbook example of sequences that
cannot be represented by a finite automaton (thus also excluding an HMM): sequences of the form
$a^nb^n$. The sequence alternates blocks of random length containing only
\texttt{a}'s and only \texttt{b}'s, with the constraint that the
length of a \texttt{b}-block is equal to the length of the
\texttt{a}-block preceding it. The blocks are separated with newlines.

\begin{Example}
\dataexstyle{aaaaaaa}
\\\dataexstyle{bbbbbbb}
\\\dataexstyle{aaaaaaaaaaaaaaaaaaaaaaaaaaaaaaaaaaaaaaaaaaaaaaaaaaaaa}
\\\dataexstyle{bbbbbbbbbbbbbbbbbbbbbbbbbbbbbbbbbbbbbbbbbbbbbbbbbbbbb}
\\\dataexstyle{aaaaaaaaaaaaaaaaaaaaaaaaaaa}
\\\dataexstyle{bbbbbbbbbbbbbbbbbbbbbbbbbbb}
\\\dataexstyle{...}
\caption{$a^nb^n$}
\label{ex:anbn}
\end{Example}

Seen as a temporal sequence, this exhibits long-term dependencies,
especially if the block lengths used in the training sequence are long.
GLNNs are found to be able to learn this model within minutes with a
training set of as few as 10 examples with the block lengths ranging
in the thousands.

\bigskip

Experiments for each of these examples are given in
Section~\ref{sec:exp}, both for GLNNs and more traditional RNNs: a GLNN
or RNN
network is trained on a single (long) training
sequence\footnote{\label{foot:singleseq}We chose
a single long training sequence rather than several short sequences,
first, to avoid giving the algorithms a hint about the time
scales at play; second, because in some of the problems presented here,
there are no marked cuts (music example), or finding the relevant cuts is
part of the game ($a^nb^n$ example); third, because having several
training sequences is not always relevant, e.g., if there is a single
temporal stream of data.} and evaluated on
an independent validation sequence, for a given computation time. More
experiments attempt to isolate
the respective contributions of the three changes introduced (leakiness,
gatedness, and Riemannian training). 
Hidden Markov models, LSTMs, and classical text compression
methods are included as a baseline.

The code for these experiments can be downloaded at
\url{http://www.yann-ollivier.org/rech/code/glnn/code_glnn_exptest.tar.gz}

\section{Definition of the models}

\subsection{Generative models for sequential data}

A generative model for symbolic sequences is a model which produces an
infinite random sequence of symbols $(x_0,\ldots,x_t,\ldots)$ over a
finite alphabet $\A$. The model depends on a set of internal parameters
$\theta$: each $\theta$ defines a probability distribution
$\Pr\nolimits_\theta((x_t)_{t=0,1,\ldots})$ over the set of infinite sequences. Given an actual
training sequence $(x_t)$, the goal of learning is to find the value of
$\theta$ that maximizes the probability of the training sequence $(x_t)$:
\begin{align}
\theta &= \argmax_\theta \Pr\nolimits_\theta((x_t)_{t=0,1,\ldots})
=\argmax_\theta \log \Pr\nolimits_\theta((x_t)_{t=0,1,\ldots})
\\&=\argmax_\theta \sum_t \log \Pr\nolimits_\theta(x_t|x_0x_1\ldots x_{t-1})
\end{align}
where the latter sum can usually be computed step by step.
This value of $\theta$ is then used for prediction of future
observations, generation of new similar sequences, or
compression of the training sequence.

The generative models considered here work in an iterative way.  At each
time step $t$, the system has an internal state. This internal state is
used to compute a probability distribution $\pi_t$ over the alphabet. The
symbol $x_t$ printed at time $t$ is drawn from this distribution $\pi_t$.
Then the new internal state as time $t+1$ is a deterministic or random
function of the internal state at time $t$ together with the symbol $x_t$
just printed.

Computing the probability of an actual training sequence $(x_t)$ can be done
iteratively, by computing the probability $\pi_0$ assigned by the model
to the first symbol $x_0$, then revealing the actual value of $x_0$,
using this $x_0$ to compute the internal state at time $1$, which is used
to compute the probabilistic distribution of $x_1$, etc\@. (\emph{forward
pass}).

In a variant of the problem, only some of the symbols in the sequence
$(x_t)$ have to be predicted, while the others are given ``for free''.
For instance, in a classification task the sequence $(x_t)$ might be of
the form $y_0z_0y_1z_1y_2z_2\ldots$ where for each instance $y_i$ we have
to predict the corresponding label $z_i$. In this case the problem is to
find the $\theta$ maximizing the probability of those symbols to be
predicted:
\begin{equation}
\label{eq:tobepredicted}
\theta=\argmax_\theta \sum_t \chi_t \log
\Pr\nolimits_\theta(x_t|x_0x_1\ldots x_{t-1})
\end{equation}
where
\begin{equation}
\chi_t=
\begin{cases}
1 & \text{if $x_t$ is to be predicted,}
\\
0 & \text{otherwise.}
\end{cases}
\end{equation}

\subsection{Recurrent neural network models}

We now present the recurrent network models discussed in this work. These
include ordinary recurrent neural networks (RNNs), gated neural
networks (GNNs), and leaky GNNs (GLNNs).

Neural network--based  models use a finite oriented graph $\network$, the
\emph{network}, over a set of \emph{units}.  The internal state  is a
real-valued function over $\network$ (the \emph{activities}), and edges
in the graph indicate which units of the network at time $t$ contribute
to the computation of the state of units at time $t+1$.

At each time step $t$, each unit $i$ in
the network $\network$ has an \emph{activation level} $a_i^t\in\R$. As
is usual for neural networks, we include a special, always-activated unit
$i=0$ with $a_0^t\equiv 1$, used to represent the so-called ``biases''.
The activation levels at time $t$ are used to compute the output of the
network at time $t$ and the activation levels at time $t+1$.
This \emph{transition function} is different for RNNs, GNNs, and GLNNs, as
defined below (Sections~\ref{sec:RNN}--\ref{sec:GLNN}).

For the output of the network we always use the softmax function:
each unit
$i\in\network$ (including $i=0$)
has time-independent \emph{writing weights} $w_{ix}$ for each symbol $x$
in the alphabet $\A$. At each time, the network outputs a random symbol
$x\in\A$ with probabilities given by the exponential of the writing
weights weighted by the activation levels at that time:
\begin{equation}
\label{eq:output}
\pi_t(x)\deq \frac{\e^{\sum_i a^t_i w_{ix}}}{\sum_{y\in\A} \e^{\sum_i
a^t_i w_{iy}}}
\end{equation}
where $\pi_t(x)$ is the probability to print $x\in\A$. This
allows any active unit to sway the result by using a large enough
weight. One effect of this ``non-linear voting'' is to easily represent
intersections of constraints: If an active unit puts high weight on a subset of
the alphabet, and another active unit puts high weight on another subset
of
the alphabet, only the symbols in the intersection of these subsets will
have high probability.

Thus, given the activities $(a^t_i)_{i\in\network}$ at time $t$, the network prints a
random symbol $x_t$ drawn from $\pi_t$. Then the network uses its current
state and its own output $x_t$ to compute the activities at time $t+1$:
the $(a^{t+1}_i)$
are a deterministic function of both the $(a^t_i)_{i\in\network}$ and
$x_t$.

Given the writing weights $w_{ix}$, the model-specific transition
function (depending on model-specific transition parameters $\tau$),
and the initial activation levels $a^0_i$, the model
produces a random sequence of symbols $x_0,x_1,\ldots,x_t,\ldots$. Given
a training sequence $(x_t)$, the goal of training is to find parameters
$w_{ix}$, $\tau$ and $a^0_i$ maximizing the probability to print
$(x_t)$:
\begin{equation}
\Pr((x_t)_{t=0,\ldots,T-1})=\prod_{t=0}^{T-1} \pi_t(x_t)
\end{equation}

The parameters $\theta=(w, \tau, a^0)$ can be trained by gradient ascent
$\theta\gets \theta+\eta\frac{\partial \log \Pr(x)}{\partial \theta}$.
The gradient of the (log-)probability to print $(x_t)$ with respect to the
parameters can be computed by the standard backpropagation
through time technique \cite{Rumelhart1987,Jaeger_tutorial}.
Appendix~\ref{sec:backprop} describes backpropagation through time
for the GLNN model (Proposition~\ref{prop:glnnder}).

However, here we will use gradient ascents in suitable,
non-trivial metrics $\norm{\theta-\theta'}$ given by a symmetric,
positive-definite matrix $M(\theta)$.  The corresponding gradient ascent
will take the form
$\theta\gets \theta+\eta M(\theta)^{-1}\frac{\partial \log
\Pr(x)}{\partial \theta}$ (see Section~\ref{sec:metricsintro}).
These metrics are built in Sections~\ref{sec:recmetric}--\ref{sec:bpm} to
achieve reparametrization invariance at a reasonable computational cost,
based on ideas from \cite{gradnn}.

We first give the full specification for the three neural network
models used.

\subsubsection{Recurrent Neural Networks}
\label{sec:RNN}

In this article we use the following transition function to compute the
RNN activation levels at step $t+1$
(see for instance \cite{Jaeger_tutorial}):
\begin{align}
\label{eq:rnn}
V^{t+1}_j &\deq
\rho_{jx_t}+\sum_i \tau_{ij} a_i^t
\\
a^{t+1}_{j} &\deq
\actf(V^{t+1}_j),
\end{align}
where $\actf$ is a fixed activation function, $x_t\in\A$ is the symbol
printed at time $t$, and the sum runs
over all edges $ij$ in the network. The sum also includes the
always-activated unit $i=0$ to represent ``biases''\footnote{Biases are actually redundant in this
case: the bias $\tau_{0i}$ at unit $i$ has the same effect as adding
$\tau_{0i}$ to all the input weights $\rho_{ix}$ for all symbols $x$,
since at any time, one and exactly one symbol is active. Still, since
backpropagation is not parametrization-invariant, using or not using these
biases has an effect on learning.}. 

The parameters to be trained are the input parameters $\rho_{jx_t}$
and the transition parameters $\tau_{ij}$. The parameter $\rho_{ix_t}$
can equivalently be thought of as a connection weight from
an input unit activated when reading $x_t$.

Two standard choices for the activation function are the logistic
function
$
\actf(V)\deq \e^V/(1+\e^V)=1/(1+\e^{-V})
$
and the hyperbolic tangent $\actf(V)\deq \tanh(V)$. Actually the two are
related: one is obtained from the other by an affine transform of $V$
and $a$. Traditional learning procedures would yield different results
for these two choices. 
With the training procedures below using an invariant metric, using the $\tanh$ function instead of the logistic function would result in the same
learning trajectory so that this choice is indifferent. To fix ideas, the experiments were implemented
using $\tanh$.

\subsubsection{Gated Neural Networks}

GNNs are an extension of recurrent neural networks, in which the neural
network transition function governing the new activations depends on the
last symbol written. This is inspired by finite automata.
Such models have also been used in
\cite{Sutskever_text2011}, the main difference being the
non-linear softmax function \eqref{eq:output} we use for the output.

In GNNs the activation levels at step $t+1$
are given by
\begin{align}
\label{eq:cnn}
V^{t+1}_j &\deq
\sum_i a^t_i \,\tau_{i j x_t}
\\
a^{t+1}_{j} &\deq
\actf(V^{t+1}_j),
\end{align}
where $\actf$ is the same activation function as in RNNs. The sum includes the
always-activated unit $i=0$. 

In the above, $x_t\in\A$ is the symbol printed at step $t$, and the parameters
$\tau_{ijx}$ are the \emph{transition weights} from unit $i$ to unit $j$
given context $x\in\A$: contrary to RNNs, $\tau_{ijx}$ depends on the
current signal $x$.
This amounts to having an RNN with different parameters $\tau$ for each
symbol $x$ in the alphabet. (This is not specific to
discrete-valued sequences here: a continuous vector-valued signal
$x_t$ with components $x^k_t$ could trigger the use of $\sum_k x_t^k
\tau_{ijk}$ as transition coefficients at time $t$.)

Hidden Markov models are GNNs with linear activation function
\cite{Bridle1990}: if we set
$\actf(V)\deq V$ and if $\tau_{ijx}$ is set to $(\text{HMM probability that
unit $i$ prints symbol $x$}) \times (\text{HMM transition probability from $i$
to $j$})$, then the GNN transition \eqref{eq:cnn} yields the update equation
for the HMM forward probabilities\footnote{More precisely $a^t_i$ becomes
the probability to have emitted
$y_0,\ldots,y_{t-1}$ and be in state $i$ at time $t$, i.e., the HMM
probabilities right before emitting $x_t$ but after the $t-1\to t$ state
transition.}. If, in addition, we replace the
softmax output \eqref{eq:output} with a linear output $
\pi_t(x)\deq \frac{{\sum_i a^t_i w_{ix}}}{{\sum_i
a^t_i }}$ where $w_{ix}$ is the HMM probability to write $x$ in state
$i$, then the GNN model exactly reduces to the HMM
model.\footnote{Conversely, any system of the form
$a^{t+1}=F(a^t,x_t)$ and $\mathrm{law}(x_{t+1})=G(a^{t+1})$, can be viewed as
a Markov process on the \emph{infinite} continuous space in which
$(a^t,x_t)$ take values.}

GNNs have more parameters than standard recurrent networks,
because each edge carries a parameter for each letter in the alphabet.
This can be a problem for very large alphabets (e.g., when each symbol
represents a word of a natural language): even storing the
parameters can become costly. This is discussed in
\cite{Sutskever_text2011}, where a factorization technique is applied to
alleviate this problem.

\subsubsection{Gated Leaky Neural Networks}
\label{sec:GLNN}

Gated leaky neural networks are a variation over GNNs which
allow for better handling of some distant temporal dependencies. They are
better understood by a detour through continuous-time models.
In GNNs we have
$V^{t+1}_j=\sum_i \tau_{ijx_t}a^t_i$. One possible way to
define a continuous-time analogue is to set
\begin{equation}
\label{eq:continuous}
\frac{\mathrm{d} V^t_j}{\mathrm{d} t}=\sum_i \tau_{ijx_t}a^t_i
\end{equation}
and set $a^t_j=\actf(V^t_j)$ as before. See \cite{Jaeger_tutorial} for
``continuous-time'' or ``leaky'' neural networks.

This produces an ``integration effect'': units become activated when a
certain signal $x_t$ occurs, and stay activated until another event
occurs. Importantly, the transition coefficient $\tau_{iix_t}$ from $i$
to $i$ itself provides a feedback control. For this reason, in our
applications, loops $i\to i$
are always included in the graph of the network.

Here, contrary to the models in \cite{Jaeger_tutorial}, the differential equation is written over
$V$ which results in a slightly different equation for the activity
$a$.\footnote{Making $V$ rather than $a$ the leaky variable comes from
looking for the simplest possible nonlinear dynamics in the context of
differential geometry for neural networks \cite{gradnn}. In full
generality, if the activity unit $j$ is a point $\mathfrak{a}_j$ in
a manifold $\mathcal{A}_j$, the continuous-time dynamics will be
$\mathrm{d}\mathfrak{a}_j/\mathrm{d} t=F_j((\mathfrak{a}_i)_{i\to j},x_t)$ where $F_j$ is
a vector field on $\mathcal{A}_j$ depending on the activitives of units
connected to $j$ and on the current signal $x_t$. Looking for dynamics
with a simple form, it makes sense to assume that the
vector-field--valued function $F_j$ is the product of a fixed vector
field $F_j^0$ times a real-valued function of the
of the $\mathfrak{a}_i$, and that the latter decomposes as a sum of the
influences of individual units $i$, namely:
$F_j((\mathfrak{a}_i)_{i\to j},x_t)=(\sum_i
f_i(\mathfrak{a}_i,x_t))F_j^0$. For one-dimensional activitives, if
$F_j^0$ does not vanish, there always exists a particular chart of
the manifold $\mathcal{A}_j$, unique up to an affine transform, in which $F_j^0$ is
constant: we call this chart $V_j$. Further assuming that
$f_i(\mathfrak{a}_i,x_t)$ decomposes as a product of a function of $x_t$
and a function of $\mathfrak{a}_i$, namely
$f_i(\mathfrak{a}_i,x_t)=\tau_i(x_t) g_i(\mathfrak{a}_i)$, we can set
$a_i\deq g_i(\mathfrak{a}_i)$, and we obtain the
dynamics \eqref{eq:continuous}. Both variables $V$ and $a$ are thus
recovered uniquely up to affine transform, respectively, as the variable
that makes the time evolution uniform and the variable that makes the
contribution of incoming units additive.}

Gated leaky neural networks are obtained by
the obvious time discretization of this evolution equation. This is
summed up in the following definition.

\begin{defi}
\label{def:glnn}
A \emph{gated leaky neural network} (GLNN) is a network as
above, subjected to the evolution equation
\begin{equation}
V^{t+1}_j\deq V^t_j+\sum_i \tau_{ijx_t}a^t_i
,\qquad a^t_j\deq \actf(V^t_j)
\end{equation}
(where the sum includes the always-activated unit $i=0$).
The probability to output symbol $x$ at time $t$ is given by
\begin{equation}
\pi_t(x)\deq \frac{\e^{\sum_i a^t_i w_{ix}}}{\sum_{y\in\A} \e^{\sum_i
a^t_i w_{iy}}}
\end{equation}
\end{defi}

Appendix~\ref{sec:init} provides a further discussion of the integrating
effect by studying the linearized regime. This is useful to gain an
intuition into GLNN behavior and to obtain a sensible parameter
initialization.

%
%
%

\section{An algorithm for GLNN training}
\label{sec:algo}

In Section~\ref{sec:theory} we expose theoretical principles along which
to build Riemannian algorithms for RNN, GNN and GLNN training.  For
convenience, we first collect here the explicit form of the final
algorithm obtained for GLNNs, and discuss its algorithmic cost.

The derivatives of the log-likelihood of the training data
with respect to the writing and transition weights,
can be computed using backpropagation through time adapted to GLNNs
(Appendix~\ref{sec:backprop}).
These derivatives are turned into a parameter update 
\begin{equation}
\theta\gets
\theta+\eta M(\theta)^{-1} \frac{\partial \log \Pr(x)}{\partial \theta}
\end{equation}
through a suitable
metric $M(\theta)$. We present two algorithmically efficient choices for
$M$: the \emph{recurrent backpropagated metric} (RBPM) and the
\emph{recurrent unitwise outer product metric} (RUOP
metric).

For the update of the writing weights $w_{ix}$, we use the quasi-diagonal
reduction \cite[Sect.~2.3]{gradnn} of the Hessian or Fisher information matrix (the
two coincide in this case) as the metric. Quasi-diagonal reduction is a
process producing an update with algorithmic cost close to using only the
diagonal of the matrix, yet has some of the reparametrization invariance
properties of the full matrix. The expression for this metric on $w_{ix}$ is worked
out in Section~\ref{sec:wmetric}.

The metric $M$ used for updating the transition weights $\tau_{ijx}$ is
built in Sections~\ref{sec:recmetric}--\ref{sec:bpm}. First, in
Section~\ref{sec:recmetric} we build a metric on recurrent networks from
any metric on feedforward networks. This involves ``time-unfolding''
\cite{Rumelhart1987,Jaeger_tutorial} the recurrent network to view it as a
feedforward network with $T$ times as many units ($T$ being the length
of the training data), and then summing the feedforward metric
over time (Definition~\ref{def:recmetric}). In Sections~\ref{sec:ruop}
and~\ref{sec:bpm} we carry out this procedure explicitly for two
feedforward metrics described in \cite{gradnn}: this yields
the RUOP metric and the RBPM, respectively.

Before starting the gradient ascent, the parameters of the network are
initialized so that at startup, the activation of each unit over time is
a random linear combination of the symbols $x_t$ observed in the recent
past.  As this latter point provides interesting insight into the
behavior of GLNNs, we discuss it in Appendix~\ref{sec:init}.

\paragraph{Algorithm description.}
Training consists in adjusting the writing weights $w_{ix}$, transition
weights $\tau_{ijx}$, and starting values $V_i^0$ (used by the network at
$t=0$), to increase the log-likelihood of the training sequence $(x_t)_t$
under the model.

The variable $\chi_t$ encodes which symbols in the sequence have
to be predicted: it is set to $1$ if the symbol $x_t$ has
to be predicted, and to $0$ if $x_t$ is given. Namely, the problem to be
solved is
\begin{equation}
\argmax_{w,\tau,V^0} \sum_t \chi_t \log \pi_t(x_t)
\end{equation}
where $\pi_t$ is the probability attributed by the network to the next
symbol knowing $x_0,\ldots,x_{t-1}$.

For simplicity we work with a single (long) training sequence
$(x_t)_{t=0,\ldots,T-1}$; the algorithm can be extended in a
straightforward manner to cover the case of several training examples, or
mini-batches of training sequences (as in a stochastic gradient
algorithm), simply by summing the gradients $W$, $G$ and the metrics
$\tilde h$, $\tilde M$ below over
the training examples.

The procedure alternates between a gradient step with respect to the
$w_{ix}$, and a gradient step with respect to the $\tau_{ijx}$ and
$V_i^0$, with two distinct learning rates $\eta_w$ and
$\eta_\tau$. We describe these two steps in turn. It is important to
start with an update of $w_{ix}$, otherwise the metric at
startup may be singular.

In the following expressions, all sums over units $i$ in the network
$\network$
include the always-activated unit $i=0$ with $a_0^t\equiv 1$.

\paragraph{Gradient update for the writing weights $w_{ix}$.} This is
done according to the following steps.
\begin{enumerate}
\item Forward pass: Compute the activations of the network over the
training sequence $(x_t)_{t=0,\ldots,T-1}$, using the GLNN evolution equations in
Definition~\ref{def:glnn}.

\item Compute the partial derivatives with respect to the writing
weights:
\begin{equation}
W_{iy}=
\sum_{t} \chi_t\,a_i^t \left(\1_{x_t=y}-\pi_t(y)\right)
\end{equation}

\item Compute the following terms of the Hessian (or Fisher information
matrix) of the log-likelihood with respect to $w$, using
\begin{align}
\label{eq:hessian}
h_{ii}^y&=\eps_y+\sum_{t=0}^{T-1} \chi_t\,(a_i^t)^2 \pi_t(y)(1-\pi_t(y))
, \qquad i\in\network,\,y\in\A
\\
h_{0i}^y&=\sum_{t=0}^{T-1} \chi_t \,a_i^t \, \pi_t(y)(1-\pi_t(y))
, \qquad i\neq 0,\,y\in\A
\end{align}
where $\eps_y$ is a dampening term to avoid divisions by $0$. We set
$\eps_y$ to the frequency of $y$ in the training sequence plus the
machine epsilon.

\item Update the weights using the quasi-diagonal reduction of the
inverse Hessian:
\begin{align}
\label{eq:qdw1}
w_{iy}\gets& w_{iy}+\eta_w\,
\frac{W_{iy}-W_{0y}h_{0i}^y/h_{00}^y}{h_{ii}^y-(h_{0i}^y)^2/h_{00}^y}
\qquad i\neq 0
\\
\label{eq:qdw2}
w_{0y}\gets& w_{0y}+\eta_w\, \left(
\frac{W_{0y}}{h_{00}^y}-\sum_{i\neq
0}\frac{h_{0i}^y}{h_{00}^y}\,\frac{W_{iy}-W_{0y}h_{0i}^y/h_{00}^y}{h_{ii}^y-(h_{0i}^y)^2/h_{00}^y}
\right)
\end{align}
(These formulas may look surprising, but they amount to using weighted
\emph{covariances} over time between desired output and activity of unit
$i$, rather than just sums over time \cite[Sect.~1.1]{gradnn}; the constant terms
are transferred to the
always-activated unit.)
\end{enumerate}

\paragraph{Gradient update for the transition weights $\tau_{ijx}$.} This
goes as follows.

\begin{enumerate}
\item Forward pass: Compute the activations of the network over the
training sequence $(x_t)_{t=0,\ldots,T-1}$, using the GLNN evolution equations in
Definition~\ref{def:glnn}.

\item Backward pass: Compute the backpropagated values
$B_i^t$ for each
unit $i\neq 0$ using
\begin{equation}
B^t_i=B^{t+1}_i+\actf'(V^t_i)
\left(
\chi_t\left(w_{ix_t}-\tsum_y \pi_t(y)w_{iy}\right)+\sum_j \tau_{i j
x_t} B^{t+1}_j
\right)
\end{equation}
initialized with $B^T_j=0$. This is the derivative of data
log-likelihood with respect to $V_i^t$. Here $\actf'$ is the derivative of the
activation function.

\item Compute the following ``modulus'' $\tilde m_i^t$ for each unit
$i\neq 0$ at each time $t$. In the RUOP variant, simply set
\begin{equation}
\tilde m_i^t=(B^t_i)^2
\end{equation}
In the RBPM variant, set by induction from $t+1$ to $t$:
\begin{equation}
\label{eq:explicitbpmod}
\begin{split}
\tilde m_i^t={}&\actf'(V_i^t)^2 
\left(
\chi_t \left(\tsum_y \pi_t(y)w_{iy}^2-(\tsum_y
\pi_t(y)w_{iy})^2\right)
+\sum_{j\neq i} \left(\tau_{ijx_t}\right)^2 \tilde m_j^{t+1}
\right)
\\&+\left(1+\tau_{iix_t}s'(V_i^t)\right)^2\tilde m_i^{t+1}
\end{split}
\end{equation}
initialized with $\tilde m_i^T=0$.

\item For each unit $j\neq 0$, for each symbol $y\in \A$, compute the
following vector $G^{(jy)}_i$ and matrix $\tilde M^{(jy)}_{ii'}$ indexed
by the units $i$ with $i\to j$ in the network $\network$, including $i,i'=0$.
\begin{equation}
G^{(jy)}_i=\sum_{t=0}^{T-1}\1_{x_t=y} \,a^t_i \, B^{t+1}_j
\end{equation}
(this is the derivative of the log-likelihood with respect to
$\tau_{ijx}$) and
\begin{equation}
\label{eq:tildeM}
\tilde M^{(jy)}_{ii'} =\sum_{t=0}^{T-1}\1_{x_t=y} \, a^t_i \,a^t_{i'}\,\tilde
m^{t+1}_j
\end{equation}
Dampen the matrix $\tilde M^{(jy)}_{ii'}$ by adding $\eps$ to the
diagonal (we use $\eps=1$ which is small compared to the $T$ terms in the
sum
making up $\tilde M$).

\item Set
\begin{equation}
G^{(jy)}\gets (\tilde M^{(jy)})^{-1} G^{(jy)}
\end{equation}
and update the transition weights with
\begin{equation}
\tau_{ijy}\gets \tau_{ijy}+\eta_\tau \, G^{(jy)}_i
\end{equation}
for each $j\neq 0$ and $y\in\A$.

\item Update the starting values $V_j^0$ with
\begin{equation}
V_j^0\gets V_j^0 +\eta_\tau \, B^0_j /(\tilde m^0_j+\eps)
\end{equation}
(this is obtained by analogy: this would be the update of $\tau_{ijy}$ with
$i=0$ and $y$ a special dummy symbol read at startup---consistently with
the fact that $\tilde m^0_j$ and $B^0_j$ have not been used to update
$\tau$).
\end{enumerate}

\paragraph{Initialization of the parameters.} At startup, the network
$\network$ is chosen as an oriented random graph with $d$ distinct edges from each unit
$i$, always including a loop $i\to i$.
For the tanh activation
function, the parameters are set up as
follows (see the derivation in Appendix~\ref{sec:init}):
\begin{equation}
\label{eq:initw}
w_{0y}\gets \log \nu_y, \qquad w_{iy}\gets 0\qquad (i\neq 0),
\end{equation}
where $\nu_y=\frac{\sum_t \chi_t \,\1_{x_t=y}}{\sum_t \chi_t}$ is the frequency of symbol $y$ among symbols to
be predicted in the training data (this
way the initial model is an i.i.d.\ sequence with the correct
frequencies). The transition parameters are set so that each unit's
activation reflects a random linear combination of the signal in some
time range, as computed in Appendix~\ref{sec:init} from the linearization
of the network dynamics, namely
\begin{equation}
\label{eq:initdiag}
\tau_{iiy}\gets -\alpha,\qquad \tau_{ijy}\gets 0 \qquad (i\neq j,i\neq 0)
\end{equation}
and
\begin{equation}
\label{eq:initread}
\tau_{0jy}\gets\beta_j+\frac{\mu_j}{4}(u_{jy}-\tsum_{y'}\tilde\nu_{y'}u_{jy'})
\end{equation}
where the $u_{jy}$ are independent random variables uniformly
distributed in $[0;1]$, $\tilde\nu_{y}=\frac{\sum_t \1_{x_t=y}}{T}$ is
the frequency of symbol $y$ in the data, and where
\begin{equation}
\mu_j=1/(j+1),\qquad
\alpha=1/2,\qquad
\beta_j=-\sqrt{\alpha(\alpha-\mu_j)}
\end{equation}
for unit $j$ ($j\geq 1$) are adjusted to control the
effective memory\footnote{In particular, any foreknowledge of the time
scale(s)
of dependencies in the sequence may be used to choose relevant values for
$\mu_j$. With our choice, from Appendix~\ref{sec:init} the time scale for unit $j$ is $O(j)$ at
startup,
though it may change freely during learning.
Multiple time scales for recurrent networks can be found in several
places, e.g., \cite{ElHihi1995,ClockworkRNN}.} of the integrating effect at unit $j$ (see
Appendix~\ref{sec:init}). These values apply to the $\tanh$ activation function.
The initial activation values are set to
$V^0_j=\actf^{-1}(\beta_j/\alpha)$ with $\actf^{-1}$ the inverse of the
activation function.

\paragraph{Learning rate control.} Gradient ascents come with a
guarantee of improvement at each step if the learning rate is small
enough. Here we test at each step whether this is the case: If an update
of the parameter decreases data log-likelihood, the update is cancelled,
the corresponding learning rate ($\eta_w$ or $\eta_\tau$) is divided by
$2$, and the update is tried again. On the contrary, if the update
improves data log-likelihood, the corresponding learning rate is
multiplied by $1.1$. This is done separately for the writing weights and
transition weights. This is a primitive, less costly form of line search%
\footnote{Experimentally, this leads to some slight oscillating
behavior when the learning rate gets past the optimal value (as is clear
for a quadratic minimum). This might be overcome by
averaging consecutive gradient steps.}%
.

At startup the value $\eta_w=\eta_\tau=1/N$ (with $N$ the number of
units) seems to work well in practice (based on the idea that if each
unit adapts its writing weights by $O(1/N)$ then the total writing
probabilities will change by $O(1)$).


\paragraph{Computational complexity.} If the network connectivity $d$
(number of edges $i\to j$ per unit $j$) is not too large, the cost of the
steps above is comparable to that of ordinary backpropagation through
time.

Let $N$ be the network size (number of units), $A$ the alphabet size, $T$
the length of the training data, and $d$ the maximum number of edges
$i\to j$ per unit $j$ in the network.

The cost of one forward pass is $O(NTd)$ for computing the activities and
$O(NTA)$ for computing the output probabilities.
The cost of computing the quantities $W_{iy}$ is $O(NTA)$ as well, as is
the cost of computing
the Hessian values $h^y$. Applying the update of $w$ costs
$O(NA)$. Thus the cost of the $w$ update is $O(NT(d+A))$.

Computing the backpropagated values $B^t_j$ costs $O(NT(d+A))$.
The cost of computing the backpropagated modulus $\tilde m^t_i$ is
identical.

The cost of computing the gradients $G_i^{(jy)}$ is $O(NTd)$ (note that
each time $t$ contributes for only one value of $y$, namely $y=x_t$, so
that there is no $A$ factor).

The costliest operation is filling the matrices $\tilde M^{(jy)}_{ii'}$.
For a fixed $j$ and $y$ this matrix is of size $d\times d$. Computing the
entries takes time $O(Td^2)$ for each $j$, hence a total cost of
$O(NTd^2)$. (Once more, each time $t$ contributes for only one value of
$y$ so that there is no $A$ factor.) Inverting the matrix has a cost of
$O(Nd^3)$: as this requires no sum over $t$, this is generally negligible
if $T\gg d$.

Thus, the overall cost (if $T\gg d$) of one gradient step is $O(NT(d^2+A))$. This suggests
using $d\approx \sqrt{A}$. In particular if $d=O(\sqrt{A})$ the overall
cost is the same as backpropagation through time.

If network connectivity is large, there is the possibility to use the
quasi-diagonal reduction of the matrices $\tilde M$, as described in
\cite[Sect.~2.3]{gradnn}. This requires computing only the terms $\tilde
M^{(jy)}_{ii'}$ with $i=i'$ or $i=0$. This removes the $d^2$ factor and
also allows for $O(d)$ inversion, as follows.

\paragraph{Non-sparse networks: quasi-diagonal reduction.} The algorithm
above must maintain a matrix of size $d\times d$ for each unit $i$, where $d$
is the number of units $j$ pointing to $i$ in the network. When $d$ is
large this is obviously costly. The \emph{quasi-diagonal reduction}
process \cite[Sect.~2.3]{gradnn} provides a procedure linear in $d$
while keeping most invariance properties of the algorithm. This
is the procedure already used for the writing weights $w_{iy}$
in~\eqref{eq:qdw1}--\eqref{eq:qdw2}. Essentially,
at each unit $j$, the signals received from units $i\to j$ are considered
to be mutually orthogonal, except for those coming from the
always-activated unit $i=0$. Thus only the terms $\tilde M_{ii}$ and
$\tilde M_{0i}$ of the matrix are used.  The update of the transition
parameters $\tau_{ijy}$ becomes as follows.

\begin{enumerate}
\item For each unit $j\in\network$ and each symbol $y\in\A$, compute the vector
$G^{(jy)}$ as before. Compute only the terms 
$\tilde M^{(jy)}_{00}$, $\tilde M^{(jy)}_{ii}$, and $\tilde
M^{(jy)}_{0i}$ of the matrix $\tilde M^{(jy)}$ in~\eqref{eq:tildeM}.
Dampen the diagonal terms $\tilde M^{(jy)}_{00}$ and $\tilde
M^{(jy)}_{ii}$ as before.

\item Update the transition weights $\tau_{ijy}$ with
\begin{align}
\tau_{ijy}\gets& \tau_{ijy}+\eta_\tau\,
\frac{G^{(jy)}_{i}-G^{(jy)}_{0}\tilde M_{0i}^{(jy)}/\tilde M_{00}^{(jy)}}{\tilde M_{ii}^{(jy)}-(\tilde M_{0i}^{(jy)})^2/\tilde M_{00}^{(jy)}}
\qquad i\neq 0
\\
\tau_{0jy}\gets& \tau_{0jy}+\eta_\tau\, \left(
\frac{G^{(jy)}_{0}}{\tilde M_{00}^{(jy)}}-\sum_{i\neq
0}\frac{\tilde M_{0i}^{(jy)}}{\tilde M_{00}^{(jy)}}\,\frac{G^{(jy)}_{i}-G^{(jy)}_{0}\tilde M_{0i}^{(jy)}/\tilde M_{00}^{(jy)}}{\tilde M_{ii}^{(jy)}-(\tilde M_{0i}^{(jy)})^2/\tilde M_{00}^{(jy)}}
\right)
\end{align}
\end{enumerate}

\section{Constructing invariant algorithms for recurrent networks}
\label{sec:theory}

We now give the main ideas behind the construction of the algorithm
above. The approach is not specific to GLNNs and is also valid for
classical recurrent networks.

\subsection{Gradients and metrics}
\label{sec:metricsintro}

Backpropagation performs a simple gradient ascent over parameter space to
train a network.  However, for GLNNs (at least), this does not work well.
One reason is that gradient ascent trajectories depend on the chosen
numerical representation of the parameters. For instance, a
non-orthogonal change of basis in parameter space will yield different
learning trajectories; yet such changes can result from simple changes in
data representation (see the introduction of \cite{gradnn}).

This is clear from the following viewpoint. Given a real-valued function
$f$ to be maximized depending on a vector-valued parameter $\theta$, the
gradient ascent update
\begin{equation}
\theta'= \theta+\eta \frac{\partial f}{\partial \theta}
\end{equation}
with learning rate $\eta$,
can be viewed, for small $\eta$, as a maximization of $f$
penalized by the change in $\theta$, namely
\begin{equation}
\theta' \approx \argmax_{\theta'}
\left\{f(\theta')-\frac1{2\eta}\norm{\theta-\theta'}^2\right\}
\end{equation}
where the equality holds up to an approximation $O(\eta^2)$ for small
$\eta$. The term $\norm{\theta-\theta'}^2$ defines a ``cost'' of changing
$\theta$.

Clearly, different ways to represent the parameter $\theta$ as a vector
will yield different costs $\norm{\theta-\theta'}^2$. For instance, a
linear change of basis for $\theta$ amounts to replacing
$\norm{\theta-\theta'}^2$ with
$\transp{(\theta-\theta')}M(\theta-\theta')$ with $M$ a symmetric,
positive-definite matrix. The associated gradient update will then be
\begin{equation}
\theta'= \theta+\eta M^{-1}\frac{\partial f}{\partial \theta}
\end{equation}
which is the general form of a gradient ascent when no privileged norm or
basis is chosen for the parameter vector $\theta$. Moreover, in general 
the matrix $M$ may depend on the current value of $\theta$, defining a
(Riemannian) \emph{metric} in which the norm of an infinitesimal change
$\theta\to\theta+\d\theta$ of the parameter $\theta$ is
\begin{equation}
\norm{\d\theta}^2=\transp{\d\theta}M(\theta)\d\theta
\end{equation}
The gradient ascent update defined by such a metric is thus
\begin{equation}
\theta'=\theta+\eta M(\theta)^{-1}\frac{\partial f}{\partial \theta}
\end{equation}

A suitable choice of $M$ can greatly improve learning, by changing the
cost of moving into various directions. 
Amari, in particular, advocated the use of the ``natural gradient'' for
learning of probabilistic models: this is a norm
$\natnorm{\theta-\theta'}^2$ which depends on the behavior of the
probability distribution represented by $\theta$, i.e., the output
probabilities of the
network, rather than on the way
$\theta$ is decomposed as a set of numbers. Thus the natural gradient provides
invariance with respect to some arbitrary design choices. (As a
consequence, learning does not depend on whether a logistic or tanh is
used as the activation function, for instance, since one can be changed
into the other by a change of variables.)

In \cite{gradnn} we introduced several metrics for feedforward neural
networks sharing this key feature of the natural gradient, at a lesser
computational cost. The main idea is to define the metric according to
what the network does, rather than the numerical values of the
parameters.  We now show how these can  be used to build invariant
metrics for recurrent networks.

\subsection{The Fisher metric on the output units and writing weights}
\label{sec:wmetric}

\paragraph{Whole-sequence Fisher metric and conditional Fisher metric.}
Metrics for neural networks first rely on choosing a metric on the output
of the network \cite{gradnn}. Here the network's output is interpreted as
a probability distribution on the sequence $(x_t)$ printed by the
network. 
Amari's natural gradient and the metrics we use are both based on
the \emph{Fisher metric} \cite{Amari2000} on the space of probability
distributions. One way to define the Fisher metric is as an infinitesimal
Kullback--Leibler divergence between two infinitesimally close
probability distributions on the same set.

For recurrent neural networks, there is a choice as to which probability
distribution should be considered. One can either view the network as
defining a probability distribution $\Pr$ over all output sequences
$(x_0,\ldots,x_t,\ldots)$, or equivalently as defining a sequence of conditional
probability distributions $\pi_t$ for the next symbol $x_t$ knowing the
previous symbols. Thus there are two ways to define a divergence on the
parameter $\theta$ based on Kullback--Leibler divergences for
finite-length output sequences $(x_t)_{0\leq t\leq T}$. 
One is
\begin{equation}
D_1(\theta,\theta')\deq
\KL{\Pr\nolimits_\theta(x_0,\ldots,x_t,\ldots)}{\Pr\nolimits_{\theta'}(x_0,\ldots,x_t,\ldots)}
\end{equation}
where $\Pr_\theta$ is the probability distribution over the set of all sequences
$(x_0,\ldots,x_t,\ldots)$ 
defined by the network
with parameter $\theta$. The other depends on the actual training sequence
$x$ and is
\begin{align}
D_2(\theta,\theta') &\deq 
\sum_t
\KL{\Pr\nolimits_\theta(x_t|x_0,\ldots,x_{t-1})}{\Pr\nolimits_{\theta'}(x_t|x_0,\ldots,x_{t-1})}
\\&=
\sum_t \KL{\pi_t}{\pi'_t}
\end{align}
where $\pi_t$ (resp.\ $\pi'_t$) is the probability distribution on the
next symbol $x_t$ defined by the network with parameter $\theta$ (resp.\
$\theta'$) knowing past observations $x_0,\ldots,x_{t-1}$.

Arguably, $D_2$ is more adapted to prediction or (online)
compression, while $D_1$ is better suited for generalization and
learning. For
instance, if the actual training sequence starts with the letter
\texttt{a}, a gradient ascent based on $D_2$ will not care how a change
of $\theta$ affects the probability of sequences starting with a
\texttt{b}.

Assuming that the training sequence $(x_t)$ has actually been sampled from
$\Pr\nolimits_\theta$ and is long enough, and under reasonable
stationarity and ergodicity assumptions
for $\Pr\nolimits_\theta$, $D_2$ should be a reasonable approximation of
$D_1$.\footnote{Indeed one has $D_1=\E_{x\sim \Pr\nolimits_\theta} \ln
\frac{\Pr\nolimits_\theta(x)}{\Pr\nolimits_{\theta'}(x)}=\E_{x\sim
\Pr\nolimits_\theta}
\ln \frac{\prod_t \Pr\nolimits_\theta(x_t|x_0\ldots
x_{t-1})}{\prod_t \Pr\nolimits_{\theta'}(x_t|x_0\ldots x_{t-1})}
=\sum_t \E_{x\sim
\Pr\nolimits_\theta}
\ln \frac{\Pr\nolimits_\theta(x_t|x_0\ldots
x_{t-1})}{\Pr\nolimits_{\theta'}(x_t|x_0\ldots x_{t-1})}
=\sum_t \E_{(x_0\ldots x_{t-1})\sim \Pr\nolimits_\theta}
\E_{x_t \sim \Pr\nolimits_\theta(x_t|x_0\ldots x_{t-1})}
\ln \frac{\Pr\nolimits_\theta(x_t|x_0\ldots
x_{t-1})}{\Pr\nolimits_{\theta'}(x_t|x_0\ldots x_{t-1})}
=\sum_t \E_{(x_0\ldots x_{t-1})\sim \Pr\nolimits_\theta}
\KL{\Pr\nolimits_\theta(x_t|x_0\ldots
x_{t-1})}{\Pr\nolimits_{\theta'}(x_t|x_0\ldots x_{t-1})}
$ so that if averaging over $t$ in the actual training sequence is a good
approximation of averaging over $t$ for a $\Pr\nolimits_\theta$-random
sequence, then $D_1$ and $D_2$ are close.} However, in a learning situation, it may not be
reasonable to assume that $(x_t)$ is a sample from $\Pr\nolimits_\theta$
until the training of $\theta$ is finished. So we do not assume that
$D_1\approx D_2$.

Algorithmically, when an actual training sequence $x$ is given,
the conditional divergence $D_2$ is much easier to work with, because it
can be computed in linear time, whereas computing $D_1$ would require
summing over all possible sequences, or using a Monte Carlo approximation
and sampling a large number
of sequences.

For these reasons, we will define a metric based on $D_2$, i.e., on the
Fisher metric on the successive individual distributions $\pi_t$.

\paragraph{Fisher metric on the output units.} At each time step, the
output of the network is a probability distribution over the alphabet
$\A$. The set of these probability distributions is naturally endowed
with the Fisher metric: the square distance between two infinitesimally close
probability distributions $\pi$ and $\pi+\d \pi$ is
\begin{align}
\natnorm{\d\pi}^2 & \deq 2\KL{\pi}{\pi+\d\pi}
\\&=\sum_{x\in\A} \frac{(\d\pi(x))^2}{\pi(x)}
=\E_{x\sim\pi} (\d\log
\pi(x))^2
\end{align}
at second order,
where $\d\log \pi(x)=\d\pi(x)/\pi(x)$ is the resulting variation of $\log
\pi(x)$.

In the networks we consider, at each step the distribution $\pi_t$ for the next symbol is
given by a softmax output
\begin{equation}
\pi_t(x)= \frac{\e^{\sum_i a^t_i w_{ix}}}{\sum_{y\in\A} \e^{\sum_i
a^t_i w_{iy}}}
\end{equation}
for each $x$ in the alphabet $\A$. Let us set 
$E_y^t\deq \sum_ia^t_i w_{iy}$, so that $\pi_t(y)=e^{E^t_y}/\sum e^{E^t_{y'}}$. 

Then the norm $\natnorm{\d\pi_t}$ of a change $\d\pi_t$ resulting from a
change $\d E^t$ in the values of $E^t$ is, by standard arguments for
exponential families, found to be
\begin{equation}
\label{eq:natnorm}
\natnorm{\d\pi_t}^2=\sum_y \pi_t(y)(\d E_y^t)^2-\sum_{y,y'}\pi_t(y)\pi_t(y')\d E_y^t\d
E_{y'}^t
\end{equation}
(see Appendix~\ref{sec:fishout}). By a property of exponential families, this is also, for any $y''$, the Hessian of $-\log \pi_t(y'')$ with respect to the
variables $E^t$. In particular, in this situation, for the parameters
$w$, the natural gradient with
learning rate $1$ coincides with the Newton method.

\paragraph{Metric over the writing coefficients.} We can now compute the
natural metric over the writing coefficients $w_{ix}$. Let $\d w_{ix}$ be
an infinitesimal change in the parameters $w_{ix}$: this creates a change
$\d\pi_t$ in the distribution $\pi_t$, for all $t$. By the discussion
above, we are interested in the quantity \begin{equation} \sum_t
\natnorm{\d\pi_t}^2 \end{equation}

Changing the writing weights $w_{ix}$ does not change the activities of
units in the network. Consequently, we have $\d E^t_y=\sum_i a^t_i \d
w_{iy}$. Thus the above yields
\begin{equation}
\sum_t \natnorm{\d\pi_t}^2=\sum_t \sum_{y,y'}\sum_{i,i'}
\pi_t(y')(\1_{y'=y''}-\pi_t(y''))a^t_ia^t_{i'}\d w_{iy} \d w_{i'y'}
\end{equation}
so that the metric $\sum_t \natnorm{\d\pi_t}^2$ over the parameters
$w_{iy}$ is given by the Fisher matrix
\begin{equation}
I_{w_{iy}w_{i'y'}}=\sum_t \pi_t(y')(\1_{y'=y''}-\pi_t(y''))a^t_ia^t_{i'}
\end{equation}
which is also, up to sign, the Hessian of the log-likelihood of the
training sequence with respect to the parameters $w$.

This is a full matrix whose inversion can be costly.
The update of the parameters $w_{iy}$ given in Section~\ref{sec:algo}
corresponds to the quasi-diagonal inverse of this metric, whose only
non-zero terms
correspond to $y=y'$ and $i=i'$ or $i=0$.
By \cite[Sect.~2.3]{gradnn}, the quasi-diagonal inverse respects
invariance under affine reparametrization of the activities of each unit.

\subsection{Invariant metrics for recurrent networks}
\label{sec:recmetric}

The natural gradient arising from the whole-network Fisher metric is algorithmically
costly to compute for neural networks (though the ``Hessian-free''
conjugate gradient method introduced in \cite{Martens2010,
Martens2011,Martens2012} approximates it). We now introduce
metrics for recurrent networks that enjoy some of the main properties of
the Fisher metric (in particular, invariance with respect to a number of
transformations of the parameters or of the activities), at a
computational cost close to that of backpropagation through time.

Any invariant metric for feedforward networks can be used to build an
invariant metric for recurrent networks, by first ``time-unfolding'' the
network as in backpropagation through time
\cite{Rumelhart1987,Jaeger_tutorial}, and then by defining the norm of a
change of parameters of the recurrent network as a sum over time of the
norms of corresponding changes of parameters at each time in the
time-unfolded network, as follows.

A recurrent neural network with $n$ units, working on an input of length
$T$, can be considered as an ordinary feedforward neural network
with $nT$ units with shared parameters
\cite{Rumelhart1987,Jaeger_tutorial}. We will refer to it as the
\emph{time-unfolded network}.  In the time-unfolded network, a unit is a
pair $(i,t)$ with $i$ a unit in the original network and $t$ a time. The
unit $(i,t)$ directly influences the units $(j,t+1)$ where $i\to j$ is an
edge of the recurrent network. We also consider the output distribution
$\pi_t$ at time $t$ as a (probability distribution--valued) output unit
of the time-unfolded network, directly influenced by all time-unfolded units $(i,t)$.

If all time-unfolded units $(i,t)$ use the same parameters $\theta_i$ as the
corresponding unit $i$ in the recurrent network, then the behaviors of the
time-unfolded and recurrent networks coincide. Thus, let us introduce
dummy time-dependent parameters $\theta_i^t$ for unit $(i,t)$ of the
time-unfolded network, and decide that the original parameter $\theta_i$
for unit $i$ in the recurrent network is a ``meta-parameter'' of the
time-unfolded network, which sets all dummy parameters to
$\theta_i^t=\theta_i$.

We are now ready to build a metric on recurrent networks from a metric
$\norm{\cdot}$ on feedforward networks.
A variation $\d\theta$ of the parameters of the recurrent network
determines a variation $\d\theta^t$ of the (dummy) parameters of the
time-unfolded network, which is an ordinary feedforward network. Thus
we can simply set
\begin{equation}
\norm{\d\theta}^2\deq \sum_t \norm{\d\theta^t}^2
\end{equation}
where
for each $t$, $\d\theta^t$ is a variation of the parameters of an
ordinary feedforward network, for which we can use the norm
$\norm{\d\theta^t}$.

If the metric used on the time-unfolded network is reparametrization-invariant, then so
will be the metric on the recurrent network (since its definition does
not use any choice of coordinates).

Using this definition for $\norm{\d\theta}$ is actually the same as
making independent gradient updates $\d\theta^t$ for each $\theta^t$ based on
the metric $\norm{\d\theta^t}$, then projecting the resulting update onto the
subspace where the value of $\theta^t$ does not depend on $t$ (where the
projection is orthogonal in the metric $\norm{\cdot}$). Equivalently,
this amounts to making independent updates for each $\theta^t$ and then finding the time-independent update $\d\theta$ minimizing $\sum_t
\norm{\d\theta^t-\d\theta}^2$.\footnote{For these equivalent interpretations,
one has to assume that there is more than one training sequence. Indeed,
for typical choices of the feedforward network metric $\norm{\d\theta^t}$, 
with only one training sequence the metric $\norm{\d\theta^t}$ on each
individual $\theta^t$ is only of rank one, hence the corresponding
update of $\theta^t$ is
ill-defined. On the other hand, even with only one training sequence, the
metric on $\d\theta$ is generally full-rank, being a sum over time of rank-one
metrics.}

Thus, we can use any of the metrics mentioned in \cite{gradnn} for
feedforward networks. Two will be of particular interest here, but other
choices are possible; in particular, in case network connectivity is
high, quasi-diagonal reduction \cite[Sect.~2.3]{gradnn}
should be used.

\begin{defi}
\label{def:recmetric}
Let $\ffnorm{\cdot}$ be a metric for feedforward networks. The \emph{recurrent
metric associated with $\ffnorm{\cdot}$} is the metric for recurrent network
parameters defined by 
\begin{equation}
\label{eq:rmetric}
\rffnorm{\d\theta}^2\deq \sum_t \ffnorm{\d\theta^t}^2
\end{equation}
namely, by summing over time
the metric $\ffnorm{\cdot}$
on the time-unfolded network.

The \emph{recurrent backpropagated metric} (RBPM) is the norm $\rbpnorm{\cdot}$
on a recurrent network associated with the backpropagated metric
$\bpnorm{\cdot}$ on the time-unfolded network.

The \emph{recurrent unitwise outer product metric} (RUOP
metric) is the norm $\ruopnorm{\cdot}$ associated with the
unitwise outer product metric $\uopnorm{\cdot}$ on the time-unfolded network.
\end{defi}

The latter two metrics are described in more detail below. Both of them are
``unitwise'' in the sense that the incoming parameters to a unit are
orthogonal to the incoming parameters to other units, so that the
incoming parameters
to different units can be treated
independently.  (Given a unit $k$ in the network, we call \emph{incoming
parameters to $k$} the parameters directly influencing the activity of
unit $k$, namely, the weights of edges leading to $k$ and the bias of
$k$.)

\begin{rem}
We shall use these metrics only for the transition parameters $\tau$ of
recurrent networks and GLNNs. For the writing parameters $w$, the
Hessian,
or equivalently the Fisher metric, is easily computed
(Section~\ref{sec:wmetric}) and there is no reason not to use it.
\end{rem}

\begin{rem}[ (Multiple training sequences)]
Definition~\ref{def:recmetric} is given for a single training sequence
$(x_t)$. In the case of multiple training sequences, one has to first
compute the metric for each sequence separately (since the time-unfolded
networks are different if the training sequences have different lengths)
and then define a metric by averaging the square norm
$\rffnorm{\d\theta}^2$ over the training dataset, as in
\cite{gradnn}. There is a choice to be made as to whether training sequences of
different lengths should be given equal weights or weights
proportional to their lengths; the relevant choice arguably depends on the
situation at hand.
\end{rem}

\begin{rem}[ (Natural metric and recurrent natural metric)]
\label{rem:natrnat}
The natural metric of a recurrent network is defined in its own right and
should not be confused with the recurrent-natural metric obtained by
applying Definition~\ref{def:recmetric} to the natural metric of the
time-unfolded network. For the natural metric of the recurrent network, the norm of a change of parameter
$\d\theta$ is the norm of the change it induces on the network outputs
$\pi_t$.
For the recurrent-natural metric, the square norm of $\d\theta$ is the sum over time $t$,
of the square norm of the change induced on the output by a change
$\d\theta^t=\d\theta$ of the dummy parameter $\theta^t$, so that the
influence of $\d\theta$ is decomposed as the sum of its influences on
each dummy parameter $\d\theta^t$ considered independently. (Still, the
influence of $\theta^t$ on the output at times $t'\geq t$ is taken into
account.) Explicitly, if $\pi_t$ is the network output at time $t$,
then the natural norm is $\natnorm{\d\theta}^2=\sum_t
\norm{\frac{\partial\pi_{t}}{\partial \theta}\d\theta}^2$ where
$\norm{\cdot}$ is the norm on the outputs $\pi_t$. Decomposing
$\frac{\partial\pi_{t}}{\partial \theta}=\sum_{t'\leq t}
\frac{\partial\pi_{t}}{\partial \rule{0pt}{1em}\theta^{t'}}$ this is
$\sum_{t} \norm{\left(\sum_{t'\leq t} \frac{\partial\pi_{t}}{\partial
\rule{0pt}{1em}\theta^{t'}}\right)\d\theta}^2$. On the other hand
the recurrent-natural norm is
$\rnatnorm{\d\theta}^2=\sum_{t'}
\natnorm{\d\theta^{t'}}^2=\sum_{t'}\sum_{t\geq t'}
\norm{\frac{\partial\pi_{t}}{\partial
\rule{0pt}{1em}\theta^{t'}}\d\theta}^2$
which is generally different and accounts for fewer cross-time dependencies.
\end{rem}

We now turn to obtaining more explicit forms of these metrics for the case
of GLNNs. We describe, in turn, 
the RUOP metric and the RBPM.
For simplicity we will assume that all symbols in the
sequence have to be predicted ($\chi_t\equiv 1$). Section~\ref{sec:algo}
includes the final formulas for the general case.

\subsection{The recurrent unitwise outer product metric}
\label{sec:ruop}

Let us now describe the recurrent unitwise outer product metric
(RUOP metric) in more detail.

We briefly recall the definition of the (non-recurrent) unitwise outer
product
metric. Suppose we have a loss function $L$ depending on a parameter
$\theta$, and moreover that $L$ decomposes as a sum or average $L=\E_{x\in\D}
\ell(x)$ of a loss function $\ell$ over individual data samples $x$ in a
dataset $\D$. The outer products of the differentials $\frac{\partial
\ell(x)}{\partial \theta}$, averaged over $x$, provide a metric on
$\theta$, namely, $\E_{x\in\D} \frac{\partial
\ell(x)}{\partial \theta}\otimes \frac{\partial
\ell(x)}{\partial \theta}$ given by the matrix
\begin{equation}
C_{ij}=\E_{x\in\D} \frac{\partial
\ell(x)}{\partial \theta_i}\frac{\partial
\ell(x)}{\partial \theta_j}
\end{equation}
This is the outer product (OP) metric on $\theta$.

The associated gradient ascent for $L$, with step size $\theta$, is thus
$\theta\gets \theta+\eta C^{-1}\frac{\partial L}{\partial \theta}$, and
this gradient direction is parametrization-invariant. (One must be
careful that scaling $L$ by a factor $\lambda$ will result in scaling
this gradient step by $1/\lambda$, which is counter-intuitive, thus
step-size for this gradient must be carefully adjusted.)

When the loss function $\ell$ is the logarithmic loss $\ell(x)=\log \Pr_\theta(y|x)$ of a
probabilistic model $\Pr_\theta(y|x)$, as is the case for feedforward
networks with $y$ the desired output for $x$, then the OP metric
$\E_{x\in\D} \frac{\partial \log \Pr_\theta(y|x)}{\partial \theta}\otimes
\frac{\partial \log \Pr_\theta(y|x)}{\partial \theta}
$ is a
well-known approximation to the Fisher metric (for the latter, $y$ would
be sampled from
the output of the network seen as a
probability distribution, instead of using only the target value of $y$). In this context it has been used for a long time
\cite{APF00,TONGA}---sometimes under the name ``natural gradient'', though it is
in fact distinct from the Fisher metric, see discussion in
\cite{BengioNG2013} and \cite{gradnn}.

The OP metric has the following unique property: For a given increment
$\delta L$ in the value of $L$, the OP gradient step is the one for which
the increment is most uniformly spread over all samples $x\in \D$, in the
sense that the variance $\Var_{x\in\D} \delta\ell(x)$ is minimal
\cite[Prop.~15]{gradnn}.

For feedforward networks, the OP metric is given by a full matrix on
parameter space. This is computationally unacceptable for large networks; a more
manageable version is the \emph{unitwise} OP metric
(UOP metric), in which the incoming parameters for each unit are made
orthogonal \cite{gradnn}. The unitwise OP metric is still invariant under reparametrization
of the activities of each unit. This decomposition is also used in
\cite{TONGA} (together with a further low-rank approximation in each block which
breaks invariance).

The \emph{recurrent} UOP metric is obtained from the UOP metric by
Definition~\ref{def:recmetric}, 
through summing over time in the
time-unfolded network. Let $i$ be a unit in the recurrent network, and
let $\theta_i$ be the set of incoming parameters to $i$. A change
$\d\theta_i$ in $\theta_i$ results in a change $\d\theta^t_i$ of all the
dummy parameters $\theta^t_i$ of units $(i,t)$ in the time-unfolded
network. The square norm of $\d\theta_i$ in the RUOP metric is, by
definition \eqref{eq:rmetric}, the sum over $t$ of the square norms of $\d\theta^t_i$ in the UOP
metric of the time-unfolded network.

For each $t$ and each unit $i$, the unitwise OP metric on the dummy
parameter $\theta_i^t$ is given by the outer product square of 
the associated change of the objective function $\log
\Pr_\theta(x)$, namely, the outer product square of $\frac{\partial \log \Pr_\theta(x)}{\partial
\theta^t_i}$. Now $\theta^t_i$ is a dummy parameter of the
time-unfolded network, and is used exactly once during computation of the
network activities, namely, only at time $t$ to compute the activity
$V^t_i$ and $a^t_i=\actf(V^t_i)$ of unit $i$. Thus we have
\begin{equation}
\frac{\partial \log \Pr_\theta(x)}{\partial
\theta^t_i}=
\frac{\partial \log
\Pr_\theta(x)}{\partial V^t_i}
\frac{\partial V^t_i}{\partial \theta^t_i}
=
B^t_i \frac{\partial V^t_i}{\partial
\theta^t_i}
\end{equation}
where the derivatives $B^t_i\deq \frac{\partial \log
\Pr_\theta(x)}{\partial V^t_i}$ are computed in the usual way by
backpropagation through time (Appendix~\ref{sec:backprop}).

The partial derivative $\frac{\partial V^t_i}{\partial \theta^t_i}$ is
readily computed from the evolution equation defining the network: for
instance, for GLNNs, the evolution equation of the time-unfolded network
(using dummy parameters) is
$V^{t}_i= V^{t-1}_i+\sum_j \tau^t_{jix_{t-1}}a^{t-1}_j$, so that the
derivative of $V^{t}_i$ w.r.t.\ the parameter $\tau^t_{jiy}$ is
$\1_{y=x_{t-1}}a_j^{t-1}$.

The unitwise OP metric for the dummy parameter $\theta^t_i$ is
given by the outer product square of $\frac{\partial \log
\Pr_\theta(x)}{\partial
\theta^t_i}$, which by the above is $(B^t_i)^2\,\frac{\partial V^t_i}{\partial
\theta^t_i}\otimes \frac{\partial V^t_i}{\partial
\theta^t_i}$. This has to be summed over time to find the recurrent UOP
metric for the true parameter $\theta_i$. So in the end, the RUOP metric
for the incoming parameters $\theta_i$ at unit $i$ is given for each
$i$ by the
matrix
\begin{equation}
\tilde M^{(i)}_{kk'}=\sum_t (B^t_i)^2 \frac{\partial V^t_i}{\partial
(\theta^t_i)_k}\frac{\partial V^t_i}{\partial
(\theta^t_i)_{k'}}
\end{equation}
where $(\theta^t_i)_k$ denotes the $k$-th component of the parameter
$\theta^t_i$, and where the derivative is with respect to the dummy
parameter $\theta^t_i$ used only at time $t$.

For GLNNs, this results in the expression given in the algorithm of
Section~\ref{sec:algo}: 
In the end, for the GLNN transition parameter $\theta=(\tau_{jiy})_{j,i,y}$,
using that $\partial V^t_i/\partial
\tau^t_{jiy}=\1_{y=x_{t-1}}a_j^{t-1}$, 
the recurrent UOP metric is
\begin{equation}
\label{eq:uoptau}
\ruopnorm{\d\theta}^2=\sum_i\sum_{j,j'}\sum_t (B^{t+1}_i)^2
a^t_j a^t_{j'}
\d\tau^t_{jix_t}\d\tau^t_{j'ix_t}
\end{equation}
The same expression holds for GNNs (but $B$ has a different expression).

\paragraph{The form of the metric.}
Thus, we find that the RUOP metric on $\tau$ is given by a
symmetric matrix with the following properties. These remarks also hold for the
other metric we use, the RBPM below.

First, different units $i$ are orthogonal (there are no cross-terms
between $\d\tau_{jix}$ and $\d\tau_{j'i'x'}$ for $i\neq i'$).

Second, for GNNs and GLNNs, different symbols $x$ are independent:
the transition parameters
$\tau_{ijx}$ and $\tau_{ijx'}$ with $x\neq x'$ are
mutually orthogonal in the RUOP metric, i.e., 
there are no cross-terms
for $x\neq x'$. This is because, at any given time $t$, only the
parameters $\tau_{jix_t}$ using the currently read symbol $x_t$
contribute to the evolution equation.  This results in a separate
matrix $\tilde M^{(ix)}$ for each pair $ix$ in the final algorithm,
reducing computational burden.

On the other hand, for RNNs with the evolution equation
$a^{t+1}_i=\actf(\rho_{ix_t}+\tsum_j \tau_{ji} a_j^t)$, there is no such block
decomposition because the transition parameters $\tau_{ij}$ have
non-trivial scalar product with all the input parameters $\rho_{ix}$ for
all $x$; thus, handling this metric would be quadratic in alphabet size.
If alphabet size is large, one solution is to restrict input to a subset
of units. Another is to use
\emph{quasi-diagonal reduction} \cite[Sect.~2.3]{gradnn} to obtain a more lightweight
but still invariant algorithm; this was tested in Section~\ref{sec:exp}.

Third, different units $j$ and $j'$ connected to the same unit $i$ are
\emph{not} independent. (In particular, the ``biases'' $\tau_{0ix}$
corresponding to the always-activated unit $j=0$, $a_j\equiv 1$ are not
orthogonal to the other transition weights.) The cross-term between
$\d\tau_{jix}$ and $\d\tau_{j'ix}$ is \begin{equation} \sum_t \1_{x_t=x}
a_j^t a_{j'}^t (B^{t+1}_i)^2 \end{equation} Besides, the derivative of
log-likelihood with respect to $\tau_{jix}$ is $\sum_t \1_{x_t=x} a_j^t
B^{t+1}_i$ (Proposition~\ref{prop:glnnder}), and the gradient step is
obtained by applying the inverse of the matrix above to this derivative.
This problem has an interesting structure. Indeed, vectors obtained as
$M^{-1}G$ where $M$ is a matrix of the form $M_{jk}=\sum_t a_j^t a_k^t
c^t$, and $G$ of the form $G_j=\sum_t a_j^t Y^t$, are weighted
least-square regression problems: $M^{-1}G$ gives the best way to write
the vector $Y^t/c^t$, seen as a function of $t$, as a linear combination
of the family $a_j^t$, seen as functions of $t$. This is the ``best-fit''
interpretation \cite[Section~3.3]{gradnn}.

Thus, using metrics of this form, each unit $i$ in the network combines
the signals from its incoming units $j$ in an optimal way to match a
desired change in activity (given by $B^t_i$) over time. The two metrics presented here,
RUOP and RBPM, differ by the choice of the weighting
$c_t$.

\begin{rem}[ (UOP metric and recurrent UOP metric)]
\label{rem:uopruop}
The recurrent
unitwise OP metric should not be confused with the unitwise OP metric
applied to the recurrent network, which is defined in its own right but
unsuitable for several reasons. For instance, with only one training
sequence $x$, the OP metric for the recurrent network is simply
$\frac{\partial \log \Pr(x)}{\partial \theta}\otimes \frac{\partial \log
\Pr(x)}{\partial \theta}$, which is a rank-$1$ matrix and thus not
invertible. On the other hand, on a single training sequence of length $T$, the
recurrent UOP metric is a sum of $T$ matrices of rank $1$. Thus for a
recurrent network, $\ruopnorm{\cdot}\neq \uopnorm{\cdot}$ in general:
one is a time sum of outer product squares, the other is the outer
product square of a
time sum. (Compare Remark~\ref{rem:natrnat}.) So the recurrent UOP metric performs an averaging of the
metric over time rather than over samples, as is expected in a recurrent
setting.

Another similar-looking metric would be the OP metric associated with the
decomposition $\log \Pr(x)=\sum_t \log \Pr(x_t|x_0,\ldots,x_{t-1})=\sum_t
\log \pi_t(x_t)$ of
the objective function. Such a decomposition gives rise to a metric
$\sum_t (\frac{\partial \log \pi_x(x_t)}{\partial \theta})^{\otimes 2}$.
This metric is generally full-rank even for a single training sequence.
The recurrent OP metric, on the other hand, is $\sum_t (\frac{\partial
\log \Pr(x)}{\partial \theta^t})^{\otimes 2}$. So while the recurrent OP
is the sum over time of the effect of the dummy time-$t$ parameter
$\theta^t$ on the objective function, the metric just introduced is the
sum over time of the effect of the parameter $\theta$ on the $t$-th
component of the objective function. These are generally different.
Computing all partial derivatives $\frac{\partial \log
\pi_x(x_t)}{\partial \theta}$ for all $t$ and $\theta$ is algorithmically
costlier, which is why we did not use this metric.
\end{rem}

\subsection{The recurrent backpropagated metric}
\label{sec:bpm}

We now work out an explicit form for the recurrent backpropagated
metric.

For a feedforward network, the backpropagated metric (BPM), introduced in
\cite{gradnn}, is defined as follows.
Given a metric on the
output units of a network (here the Fisher metric on $\pi_t$), one can inductively
define a metric on every unit by defining the square norm $\bpnorm{\d
a_i}^2$ of a change of activity $\d a_i$ at unit $i$, as the sum
$\sum_{j,\,i\to j}
\bpnorm{\d a_j}^2$ of the square norms of the resulting changes in
activity at units $j$ \emph{directly influenced by $i$}, thus
``backpropagating'' the definition of the metric from output units to
inner units. The metric
$\bpnorm{\d a_j}^2$ at unit $j$ is then turned into a metric on the
incoming parameters to $j$, by setting $\bpnorm{\d \theta_j}^2\deq
\bpnorm{\d a_j}^2$ with $\d a_j$ the change of $a_j$ resulting from the
change $\d\theta_j$.

The \emph{recurrent} BPM is obtained from the BPM by
Definition~\ref{def:recmetric}, 
through summing over time in the 
time-unfolded network. Let $i$ be a unit in the recurrent network, and
let $\theta_i$ be the set of incoming parameters to $i$. A change
$\d\theta_i$ in $\theta_i$ results in a change $\d\theta^t_i$ of all the
dummy parameters $\theta^t_i$ of units $(i,t)$ in the time-unfolded
network. The square norm of $\d\theta_i$ in the RBPM is, by
definition \eqref{eq:rmetric}, the sum over $t$ of the square norms of
$\d\theta^t_i$ in the backpropagated metric
metric of the time-unfolded network.

So let us work out the backpropagated metric in the time-unfolded
network. The time-unfolded unit $(i,t)$ directly influences the
time-unfolded units $(j,t+1)$ for all edges $i\to j$ in the graph of the
original network, and it also directly influences the distribution
$\pi_t$ at time $t$.

Thus, let $\d a_i^t$ be an infinitesimal change in the activity of
time-unfolded unit $(i,t)$. Let $\d \pi_t$ be the resulting change in the
probability distribution $\pi_t$, and $\d a_j^{t+1}=\frac{\partial
a_j^{t+1}}{\partial a_i^t} \d a_i^t$ the resulting change in the activity of
time-unfolded unit $(j,t+1)$. The BPM is obtained by backwards induction over $t$
\begin{equation}
\label{eq:bpm}
\bpnorm{\d a_i^t}^2\deq \natnorm{\d\pi_t}^2+\sum_j \bpnorm{\d
a_j^{t+1}}^2
\end{equation}

The term $\natnorm{\d\pi_t}^2$ is readily computed from
Section~\ref{sec:wmetric}: 
in the notation above, the change in $E_y^t=\sum_j w_{jy} a_j^t$ from a change of activity in
$a_i^t$ is $\d E_y^t=w_{iy}\d a^t_i$, so that \eqref{eq:natnorm} yields
\begin{equation}
\natnorm{\d\pi_t}^2=(\d a^t_i)^2
\left(\tsum_y \pi_t(y)w_{iy}^2-(\tsum_y
\pi_t(y)w_{iy})^2\right)
\end{equation}
i.e., proportional to the $\pi_t$-variance of $w_{iy}$ (in line with the fact that
translating weights does not change output).

Since activities are one-dimensional, the backpropagated metric is simply
proportional to $\left(\d a_i^t\right)^2$, so that we have
\begin{equation}
\bpnorm{\d
a_i^t}^2 \eqd m_i^t  \left(\d a_i^t\right)^2
\end{equation}
for some positive number $m_i^t$, the
\emph{backpropagated modulus} \cite{gradnn}. The definition
\eqref{eq:bpm} of the backpropagated metric thus
translates as
\begin{equation}
m_i^t=\left(\tsum_y \pi_t(y)w_{iy}^2-(\tsum_y
\pi_t(y)w_{iy})^2\right)
+\sum_j \left(\frac{\partial
a_j^{t+1}}{\partial a_i^t}\right)^2 m_j^{t+1}
\end{equation}
(initialized with $m^T_i=0$),
in which one recognizes a source term from the output at time $t$, and a term
transmitted from $t+1$ to $t$.

It is advisable to express the
backpropagated metric using the variable $V^t_i$ rather than $a^t_i$
(because the expression for $\frac{\partial V_j^{t+1}}{\partial V_i^t}$
is simpler). The variables $V$ and $a$ correspond bijectively to each
other, and their variations are related by $\d a_i^t=\actf'(V_i^t) \d
V_i^t$ so that $\bpnorm{\d a_i^t}=m_i^t\left(\d a_i^t\right)^2=\tilde
m_i^t \left(\d
V_i^t\right)^2$ with
\begin{equation}
\tilde m_i^t\deq m^i_t\, \actf'(V_i^t)^2
\end{equation}
from which we derive the induction equation for $\tilde m$, namely
\begin{equation}
\tilde m_i^t=\actf'(V_i^t)^2 \left(\tsum_y \pi_t(y)w_{iy}^2-(\tsum_y
\pi_t(y)w_{iy})^2\right)
+\sum_j \left(\frac{\partial
V_j^{t+1}}{\partial V_i^t}\right)^2 \tilde m_j^{t+1}
\end{equation}
in which we can now easily compute the $\frac{\partial
V_j^{t+1}}{\partial V_i^t}$ term from the evolution equation defining the
recurrent network. 

For instance, for GLNNs
we have $V_j^{t+1}=V_j^t+\sum_i
\tau_{ijx_t} \actf(V_i^t)$ so we find
\begin{equation}
\frac{\partial
V_j^{t+1}}{\partial V_i^t}=\1_{i=j}+\tau_{ijx_t}\actf'(V_i^t)
\end{equation}
which, plugged into the above, yields the explicit equation
\eqref{eq:explicitbpmod} given in the algorithm description.

Once the backpropagated modulus is known, the backpropagated metric on
the dummy parameters $\theta^t_i$ at each unit $(i,t)$ of the
time-unfolded network is obtained by $\bpnorm{\d\theta^t_i}\deq\bpnorm{\d
a_i^t}$ where $\d a_i^t=\frac{\partial a^t_i}{\partial
\theta^t_i}.\d\theta^t_i$ is the variation of $a_i^t$ resulting from a
variation $\d \theta^t_i$. Thus
\begin{equation}
\bpnorm{\d\theta^t_i}^2=m^t_i\left(\frac{\partial a^t_i}{\partial
\theta^t_i}.\d\theta^t_i\right)^2=\tilde m^t_i \left(\frac{\partial V^t_i}{\partial
\theta^t_i}.\d\theta^t_i\right)^2
\end{equation}
where, as in the case of the RUOP metric above, the derivative
$\frac{\partial V^t_i}{\partial
\theta^t_i}$ can be obtained from the evolution equation defining the
network. In components, 
$\bpnorm{\d\theta^t_i}^2$ is thus given by a matrix whose $kk'$ entry is
\begin{equation}
\tilde m^t_i \frac{\partial V^t_i}{\partial
(\theta^t_i)_k}\frac{\partial V^t_i}{\partial
(\theta^t_i)_{k'}}
\end{equation}
where $(\theta^t_i)_k$ denotes the $k$-th component of the
incoming parameter
$\theta^t_i$ to unit $i$.

A parameter $\theta_i$ of the recurrent network influences all dummy
parameters $\theta^t_i$ for all $t$.
The recurrent backpropagated metric is obtained by summing the
backpropagated metric over time as in \eqref{eq:rmetric}. So in the end
the recurrent backpropagated metric for the incoming parameter
$\theta_i$ to unit $i$ is given by the matrix
\begin{equation}
\tilde M^{(i)}_{kk'}=\sum_t \tilde m^t_i \frac{\partial V^t_i}{\partial
(\theta^t_i)_k}\frac{\partial V^t_i}{\partial
(\theta^t_i)_{k'}}
\end{equation}
with $(\theta^t_i)_k$ the $k$-th component of $\theta^t_i$, and where the derivative is with respect to the dummy
parameter $\theta^t_i$ used only at time $t$.

For instance, in GLNNs, the incoming parameter to unit $i$ is
$\theta_i=(\tau_{jiy})_{j,y}$. The evolution equation
$V^{t}_i=V^{t-1}_i+\sum_j \tau^t_{jix_{t-1}} a_j^{t-1}$ using
the dummy parameters yields $\frac{\partial V^{t}_j}{\partial
\tau^t_{jiy}}=\1_{x_{t-1}=y} a^{t-1}_j$. This results in the expression
given in the algorithm of Section~\ref{sec:algo}.
In the end, for the GLNN parameter $\theta=(\tau_{jiy})_{j,i,y}$,
the recurrent backpropagated metric is
\begin{equation}
\label{eq:bptau}
\rbpnorm{\d\theta}^2=\sum_i\sum_{j,j'}\sum_t \tilde m^{t+1}_i
a^t_j a^t_{j'}
\d\tau_{jix_t}\d\tau_{j'ix_t}
\end{equation}

The structure of this metric is the same as for the RUOP metric above,
and the same remarks apply (see Section~\ref{sec:ruop}): incoming parameters to distinct units $i$ are
independent; parameters corresponding to distinct symbols $y\neq y'$ are
independent for GNNs and GLNNs but not for RNNs; finally, the transition
parameters from different units $j$ and $j'$ incoming to the same unit
$i$ are not independent, and the gradient ascent in this metric realizes,
at each unit $i$, a weighted least-square regression on the incoming signals
from units $j$ to best match a desired activation profile given
by the backpropagation values.

\subsection{Invariance of the algorithms}

Amari \cite{Amari1998,Amari2000} pioneered the use of ``invariant''
algorithms for statistical learning that do not depend on a chosen numerical
representation (parametrization) of the parameter space of the model.
Invariance can often improve performance; for instance, in the standard
RNNs in the experiments below, replacing the standard inverse diagonal
Hessian with the (invariant) quasi-diagonal inverse brings performance of
RNNs closer to that of GLNNs, at very little computational cost.

The gradient ascent presented above is invariant by reparametrization of
the activities and by reparametrization of the incoming parameters to
each unit (but not by reparametrizations mixing incoming parameters to
different units, as the natural gradient is).

This stems from its construction using a metric which depends only on the
behavior of the network. For instance, using tanh instead of sigmoid
activation function and following the same procedure would result in an
algorithm with identical learning trajectories.

However, in practice three factors limit this invariance.
\begin{enumerate}
\item The invariance holds, in theory, only for the continuous-time
gradient trajectories. The actual gradient steps with non-zero learning
rate are only approximately invariant when the learning rate is small.
Still, the actual gradient steps are exactly invariant under
\emph{affine} reparametrizations of the parameters and activity (such as
changing sigmoid into tanh).
\item Parameter initialization is done by setting numerical
values for the parameters in an explicit numerical representation.
Changing parametrization obviously means changing the initial values in
the same way. If initialization is based on an intended
parametrization-independent behavior at startup, as in
Section~\ref{sec:algo}, this is not a problem.
\item The dampening procedure for matrix inversion (the various $\eps$ terms in
Section~\ref{sec:algo}) formally breaks invariance. Using a Moore-Penrose
pseudoinverse
(which is simply the limit $\eps\to 0$) does not solve this
problem.
It would be nice to have a dampening scheme preserving
invariance\footnote{Here is a possibility for defining a matrix for the incoming parameters to
a unit $i$, which could be used as dampening the metric at $i$ in an
invariant way: 
Compute a copy of the metric (RUOP or RBPM) but replacing the actual
training sequence $(x_t)$ with a randomly generated sequence (e.g.,
uniform, or a perturbation of $(x_t)$). More copies with more random
sequences can be used until one gets a non-degenerate metric. The
resulting metric can be multiplied by a small number and used as a
dampening term. But this does not solve all problems: for
instance, if a unit $i$ has no effect whatsoever on the output given the
current parameters, the corresponding metric will vanish. It seems
difficult to define a non-zero invariant metric in the latter situation.
%
}.
\end{enumerate}

\section{Preliminary experiments}
\label{sec:exp}

Here we report a comparison of the performance of GLNNs and more
traditional RNNs on some synthetic data examples: the ``alphabet with
insertion'' (Example~\ref{ex:insert} from the Introduction), synthetic
music (Example~\ref{ex:mus}),
the distant XOR problem
(Example~\ref{ex:xor}), and finally 
the $a^nb^n$
problem
(Example~\ref{ex:anbn}). LSTMs are used as an additional benchmark.

GLNNs were trained with either the recurrent backpropagated metric or the
recurrent unitwise outer product metric, as described in Section~\ref{sec:algo}.

The reference RNN was trained using traditional (but not naive) techniques as described
below.  For the distant XOR example, RNN performance is known to
be poor unless the ``Hessian-free'' technique is used \cite{Martens2011},
so we did not test RNN on this example and instead directly compare
performance to \cite{Martens2011}.

\paragraph{Reference RNN training.} The RNN used as a baseline
is described in Section~\ref{sec:RNN}.
In particular, both this RNN and GLNNs use a softmax \eqref{eq:output} for
the probability to produce a symbol $x$ given the internal state of the
network.

RNN training is done via backpropagation through time. As plain
backpropagation was too slow, for the
parameters $w_{iy}$ the inverse diagonal Hessian (obtained from~\eqref{eq:hessian}) is applied to the gradient
update, and the learning rate for each $\rho_{ix}$ is inversely
proportional to the frequency
of symbol $x$ in the data (thus compensating for the number of terms
making up the corresponding gradient, so that rare symbols learn as fast as frequent
symbols\footnote{If $\rho_{ix}$ is seen as the weight from an input unit
activated when symbol $x$ occurs, then this is equivalent to scaling the
input unit signals to a given $L^2$ norm over time.}). A method similar
to RMSprop or Adagrad \cite{AdaGrad2011}, in which the learning rate for
each transition parameter is divided by the root mean square (RMS) average over time of
the gradient magnitude, is also reported in Table~\ref{fig:res}.

Initialization of the RNN parameters has been set along the same
principles as for GLNNs,
namely
\begin{equation}
w_{0y}\gets \log \nu_y,\qquad
\tau_{ii}\gets 1-1/i ,\qquad
\rho_{jy}\gets\frac{1}{2}(u_{jy}-\tsum_{y'}\tilde\nu_{y'}u_{jy'})
\end{equation}
with $u$ and $\tilde\nu$ as in Section~\ref{sec:algo}, and
with all other weights set to $0$,
where the symbol frequencies $\nu_y$ and $\tilde \nu_y$ are as in
Section~\ref{sec:algo},
and
the $u_{jy}$ are independent random variables uniformly
distributed in $[0;1]$. This way, at startup the activation of each unit
is given by a random linear combination of past symbols with weights
exponentially decreasing with time, with unit $i$ having a
decay time of order $i$ thanks to $\tau_{ii}$.

More combinations of models (RNN, GNN, GLNN) and training methods are
reported in Table~\ref{fig:res}.

\paragraph{LSTMs.} LSTMs \cite{HochreiterSchmidhuber1997} are included as
an additional benchmark. For this we have kept the same overall procedure
and simply replaced each RNN cell with an LSTM cell following
Eqs.~(7)--(11) in \cite{Graves2013}, and modified the gradient
accordingly. We kept the softmax output from the other models (also as
used in \cite{Graves2013}). The weights were initialized to uniform
random values in $[-0.1,0.1]$ \cite{Graves_transduction,Gers2003LSTM}. Network construction, network
sizes, and CPU time budget were identical to the other models, as
described below. Since plain gradient resulted in slow training, we have
also included a variant described above for RNNs: using the
diagonal Hessian for the writing parameters $w$, and frequency-adjusted
learning rates for the input symbols (equivalent to rescaling the
inputs). Still, training is quite slow and from Table~\ref{fig:res} it
appears that LSTMs are not competitive in this setup\footnote{Good
performance of LSTMs has been reported for one of the problems we use,
the $a^nb^n$ problem \cite{Gers2003LSTM}.
However this involved more
samples and small values of $n$ in the training set. With these
settings we were able to obtain similar results.}, at least for the
computational time budget used here.

\paragraph{Regularization.} When working with discrete alphabets, the problem arises of having
probability $0$ for certain symbols in certain situations after training;
if the trained model is used on a validation set, validation
log-likelihood can thus be very low. In our situation this is especially
the case near the beginning of the sequence: since the model is trained
on only one training sequence and has parameters for the activities at
startup, it can frequently learn to start in a specific configuration to
reproduce the first few letters of the training sequence. For this
reason, a crude regularization procedure was used: before computing log-likelihood of the validation sequence, the
prediction $\pi_t$ for the next symbol at time $t$ was replaced with
$(1-\frac{1}{t+2})\pi_t+\frac{1}{t+2}\mathrm{unif}_\A$ with
$\mathrm{unif}_\A$ the uniform distribution over the alphabet. (This kind
of regularization has some backing from information theory.)

\paragraph{Experimental setup.} The same overall procedure (construction of a random graph, learning rate
control) has been used for both GLNNs and RNNs as described in
Section~\ref{sec:algo}, following nearly identical implementations.

In each case, a single\footnote{see footnote~\ref{foot:singleseq}} training sequence
$(x_t)$ is generated using the exact synthetic model. Another,
independent sequence $(x'_t)$ is used for validation: we report the
log-likelihood (in base $2$) of the validation sequence $(x'_t)$ using
the GLNN or RNN trained on $(x_t)$. The baseline for performance is the
number of random bits used by the exact synthetic model to generate
$(x'_t)$.

As a sanity check, we also report the performance of a well-known,
efficient online text prediction method, \emph{context tree weighting}
(CTW): the algorithm is presented with the concatenation of the training and
validation sequence, and we report the number of bits used to predict the
validation sequence after having read the training sequence.

The comparison between GLNNs and RNNs is made for identical computation
time on the same machine, for a series of hyper-parameter settings
(network size and connectivity). Indeed, as RNNs and GLNNs have
different parameter sets, direct comparison for the same
network size is difficult.
Spanning different network sizes shows the performance each model can attain for a given
time budget if the right hyper-parameters are used.

In each case, the size of the network was chosen to increase from 4
units to a maximum of 256 or 512 units by increments of a factor
$\sqrt{2}$. For each network size, we tested both a sparse network with
connectivity $d=3$ edges per unit (including a loop at each unit), and a
``semi-sparse'' network with connectivity $d=\sqrt{2\#\A}$ for GLNNs and
$d=\#\A$ for RNNs, where $\#\A$ is the alphabet size; this latter choice
balances the various contributions to algorithmic complexity (see
Section~\ref{sec:algo}). This way, RNNs can take advantage of their
lesser computational sensitivity to
connectivity $d$.

For each hyper-parameter setting, the corresponding model was allowed to
learn for the same time (10 or 30 minutes depending on the example).

The experiments were run on a standard laptop computer with an Intel
Core i7-3720QM CPU at 2.60GHz \footnote{For technical reasons the
experiments for LSTMs and the RMS variant of RNNs were done on a slightly
faster machine; an empirically adjusted scaling factor was applied to the
corresponding CPU time.}, using a straightforward implementation in
C++.

The code for these experiments can be downloaded at
\url{http://www.yann-ollivier.org/rech/code/glnn/code_glnn_exptest.tar.gz}

\bigskip

Let us now discuss each example in turn.

\paragraph{Alphabet with insertions.} The synthetic generative model is
as follows. The training sequence is the concatenation of 1000 lines,
separated by a newline symbol. Each line is obtained by writing the 26
lowercase letters of the Latin alphabet, in the standard order, and then
inserting (independently) a sub-block after each letter with probability
$1/26$ for each letter. A sub-block 
starts with an opening parenthesis, followed by the
10 digits from 0 to 9 (in that order), and ends with a closing
parenthesis. After each digit in the sub-block, with probability $1/5$ a
sub-sub-block is inserted, which consists of an opening square bracket,
nine random uppercase letters chosen from A--Z, and a closing bracket.
Thus a typical line might look like\\
\dataexstyle{ab(0123[WZPYCPEEH]456789[HYDVTWATR])cdefghijklmnopqrstuvwxyz}

The validation sequence has the same law: the concatenation of 1000
independent such lines. Randomization of the innermost blocks
prevents rote learning.

\begin{figure}
\begin{centering}
\includegraphics{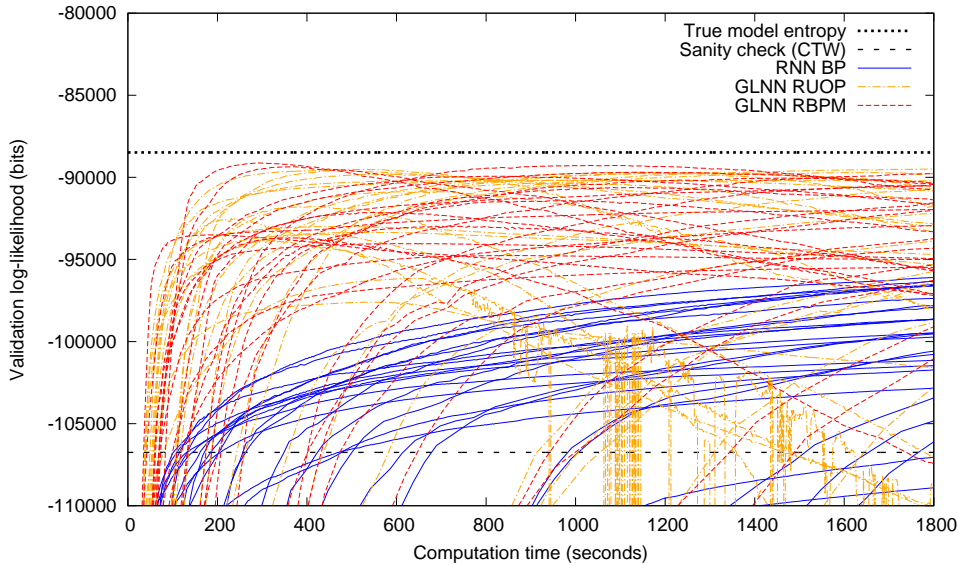}
\caption{Validation log-likelihood on the alphabet with insertions example.}
\label{fig:insert}
\end{centering}
\end{figure}

GLNNs and RNNs with a variety of network sizes ranging from 4 to 512
units, as described above, were run for 30 minutes each on the training
sequence. The validation sequence log-likelihood is reported in
Figure~\ref{fig:insert} and Table~\ref{fig:res}.

GLNNs come more than ten times closer to the true model log-likelihood
than RNNs: the best validation log-likelihood for GLNNs is -89,126 bits
while that for RNNs is -96,099 bits, compared to -88,482 bits for the
true model. Such a difference of roughly 7,000 bits represents roughly 7 bits per
line of the training sequence. Note that the cost of representing a
letter in the alphabet is $\log(26)/\log(2)\approx 4.7$ bits: this would
be the log-likelihood difference, for each line of the training sequence,
between a model that resumes at the correct place in the alphabet after a
sub-block insertion, and one that resumes at a random letter.

This is confirmed by visual inspection of the models obtained after
training. Indeed, since we train generative models, the trained network
can be used to generate new sequences, hopefully similar to the training
sequence. Doing so with RNNs and GLNNs reveals qualitative differences in
the models learned, in line with the difference in performance: After
a sub-block has been inserted, GLNNs resume at the correct letter or
sometimes one letter off the correct position in the alphabet; on the
other hand, RNNs seldom resume at the correct position.

The remaining small difference in log-likelihood between GLNNs and the true
model can, from visual inspection, be attributed to various factors:
residual errors like occasional duplicated or omitted letters, or resuming
one letter off after an insertion, as well as arguably good generalizations of
the training sequence such as having more than one sub-block between two
letters or starting a new line with a sub-block.

There is no obvious pattern of dissimilar performance between sparse and
semi-sparse networks.

However, GLNNs are apparently quite sensitive to overfitting over time:
validation log-likelihood increases at first, then steadily decreases as
parameter optimization progresses. This phenomenon is also present to a
lesser extent for RNNs, but only after much longer training times. Note
that for a given network size, GLNNs have more parameters (because each
edge has as many parameters as symbols in the alphabet $\A$).

This illustrates the importance of using a validation sequence to
stop training of GLNNs.

One GLNN run exhibits wild variations of validation log-likelihood, for
unknown reasons (perhaps a badly invertible matrix $\tilde M$).

On the other hand, surprisingly, GLNNs are less sensitive to overfitting
due to a too large network size: while increasing network size past some
value results in worse performance for RNNs (lower curves on
Figure~\ref{fig:insert}), for GLNNs it seems that the best validation
log-likelihood over an optimization trajectory stays the same for a wide
range of network sizes.

Running RNNs for longer times only partially bridges the gap in
performance:
RNNs after 4 hours are still seven times farther from the
true model than GLNNs are after 30 minutes (with a gain of 2,810 bits
of log-likelihood for RNNs).
After some time, RNNs slow down considerably or sometimes
exhibit the
same overfitting phenomenon as GLNNs and their validation performance
decreases.

Overall, the ``resume-after-insertion'' phenomenon illustrated by this
example is well captured by GLNNs.

\paragraph{Synthetic music.} The next example is synthetic music
notation, meant to illustrate the intersection of several independent
constraints. The training sequence is a succession of musical bars.
Successive musical bars are separated by a \texttt{|} symbol and a
newline symbol. Each bar is a succession of notes separated by spaces,
where each note is made of a pitch (\texttt{a},\texttt{b},\texttt{c},...) and value (\texttt{4} for a quarter note,
\texttt{2} for a half note, \texttt{4.}\ for a dotted quarter note,
etc.). In each bar, a hidden variable determines a \emph{harmony} with
three possible values I, IV, or V. If the harmony is I, every pitch in the
bar is taken uniformly at random from the set (``chord'')
\{\texttt{c},\texttt{e},\texttt{g}\}; pitches are taken from
\{\texttt{c},\texttt{f},\texttt{a}\} if harmony is IV, and from
\{\texttt{g},\texttt{b},\texttt{d}\} if harmony is V.
Harmonies in successive
bars follow a specific deterministic pattern: an 8-bar-long cycle
I-IV-I-V-I-IV-V-I as encountered in simple tunes.
Finally, in each bar, the successive durations are taken from a
finite set of 5 rhythmic possibilities (commonly encountered in
waltzes), namely: 4-4-4; 2-4; 4.-8-4; 2.; 4-4-8-8. Rhythm is chosen independently from pitch and
harmony. See Example~\ref{ex:mus}.

The training sequence is made of 2,700 musical bars.
%
The validation sequence is taken independently with the same law.

\begin{figure}
\begin{centering}
\includegraphics{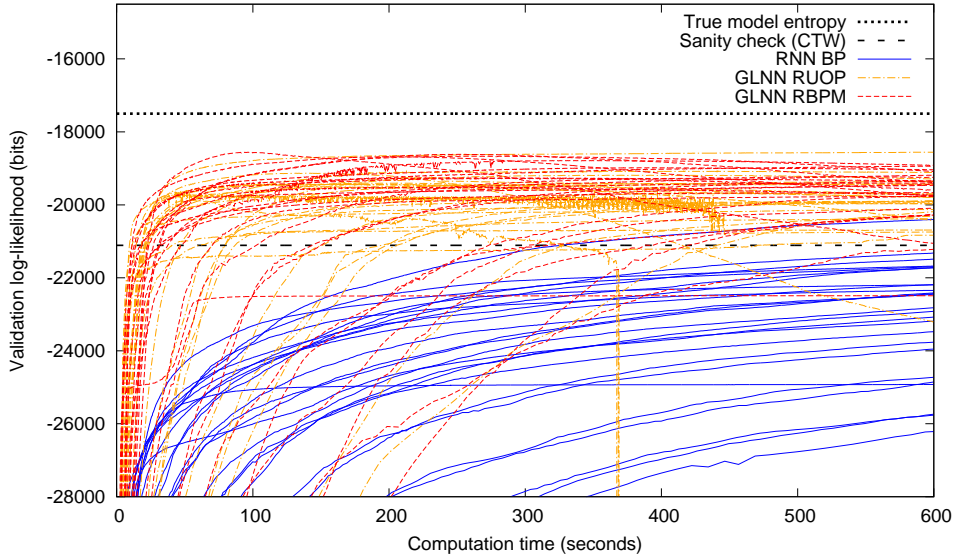}
\caption{Validation log-likelihood on the synthetic music example.}
\label{fig:mus}
\end{centering}
\end{figure}

GLNNs and RNNs with a variety of network sizes ranging from 4 to 256
units, as described above, were run for 10 minutes each on the training
sequence. The validation sequence log-likelihood is reported in
Figure~\ref{fig:mus} and Table~\ref{fig:res}.

Only one RNN run beats the sanity check (CTW).
There is a difference of roughly 2,000 bits between the best RNN and best
GLNN performance; GLNNs come roughly three times closer to the true model.

Visual inspection of the output of
the networks seen as generative models confirms that this difference is
semantically significant: GLNNs correctly learn the rhythmic and harmonic
constraints inside each bar, whereas RNNs still display ``mistakes''.

On the other hand, even GLNNs were not able to learn the underlying
8-bar-long harmonic progression, which was apparently approximated by
probabilistic transitions. This is reflected in the remaining gap between
the true model and GLNNs.

Running an RNN with backpropagation for a longer time (3 hours instead of
10 minutes) only partially bridged the gap, only bringing RNN an
additional 604 bits in log-likelihood. Once more, visual inspection of
RNN output revealed a correct learning of the possible set of rhythms,
but imperfect learning of the harmonic constraints even inside each musical bar.

The pattern of decrease in validation log-likelihood because of
overfitting is present but less pronounced than for the
alphabet-with-insertions example. Still, on Figure~\ref{fig:mus} one can
notice one GLNN run exhibiting a wild variation of validation
log-likelihood at some point.
Once more this points out the importance of using validation sets during
GLNN training, although using only one training sequence of relatively
small size may also play a role here.

\paragraph{Distant XOR.} The setting is taken from \cite{Martens2011},
after \cite{HochreiterSchmidhuber1997}; here we recast it in a symbolic
sequence setting. A parameter $T$ is fixed ($T=100$ below), which
determines the length of the instances. The training
sequence is a concatenation of lines separated by newline symbols. Each
line is
made of $T'$ random bits preceded by whitespaces, where $T'$ is taken at
random between $T$ and $1.1T$. Two of these random bits are preceded by a
special symbol \dataexstyle{X} instead of a whitespace. The positions of
these two special symbols are taken at random from the intervals
$\intint{0;T'/10}$ and $\intint{T'/10;T'/2}$ respectively. At the end of
each line, a symbol \dataexstyle{=} is inserted and is followed by a bit
giving the XOR result of the two bits following the two \dataexstyle{X}
symbols. Example~\ref{ex:xor} gives a typical training sequence.

The goal is to correctly predict the value of the final bit of each
line. So in the gradient
computation an error term is included only for the bits to be predicted,
as in
\cite{HochreiterSchmidhuber1997}. Namely, in the notation of
Section~\ref{sec:algo}, we set $\chi_t=1$ if and only if $x_{t-1}$ is the
symbol \dataexstyle{=}.

For this problem, we did not run the reference RNN and directly compared
to the best performance we found in the literature, in
\cite{Martens2011}, using ``Hessian-free'' second-order RNN training. The
success rate reported in the latter, for $T=100$, is about $25\%$
(proportion of runs achieving a classification error below $1\%$ using at
most 50,000 minibatches of 1,000 instances each).

We ran eight distinct instances of the problem, each with a different
random
training and validation sequence. Each such sequence was the
concatenation of 10,000 lines as above with $T=100$. We used a fully connected
network with 10 units.  Optimization was run for 1,500
gradient passes over the training sequence (amounting to roughly $12$
hours of computation and 750
gradient steps for each of the writing and transition parameters, since we
alternate those). We discuss the results for training using the recurrent
BPM; the results using the recurrent UOP metric are extremely similar.

\begin{figure}
\begin{centering}
\includegraphics{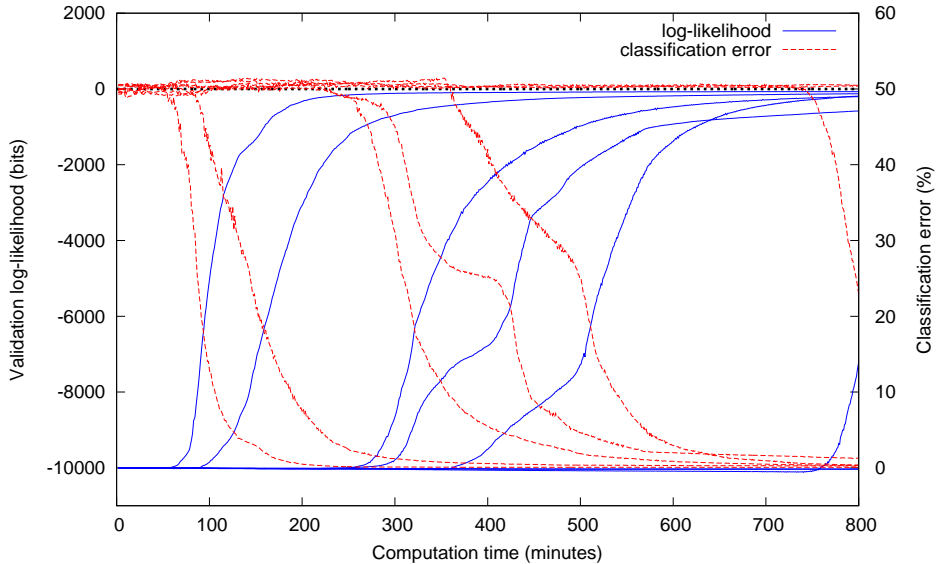}
\caption{Validation log-likelihood and classification error on the
distant XOR problem using GLNNs.}
\label{fig:xor}
\end{centering}
\end{figure}

Figure~\ref{fig:xor} reports two measures of performance on the
validation sequence: the log-likelihood score for prediction
of the final bit of each line (following the
score~\eqref{eq:tobepredicted}), and the classification error (equal to
$0$ if the correct bit value is given a probability $>1/2$ and to $1$
otherwise---this is always bounded by the log-likelihood error) expressed
as a percentage.

The results are binary: each run either successfully achieves low
error rates after enough time, or does not perform better than random
prediction.

4 out of 8 independent runs reached error rates below $1\%$ within less
than 1,500 gradient passes over the training set, and 6 out of 8 within
2,000 gradient passes. The sample size is too small to tell for sure that
this is better than the success rate in \cite{Martens2011}.  Still, the
algorithm is simpler and uses fewer training examples.

Direct comparison of the algorithmic cost with the approach in
\cite{Martens2011} is difficult, because for each gradient pass the
latter can perform up to 300 passes of the conjugate gradient algorithm
used in the implicit Hessian computation. For reference, in our approach,
each run of the experiment above (1,500 gradient passes on a training
sequence of 10,000 lines) takes slightly above 4h of CPU time on an Intel
Core i7-3720QM CPU at 2.60GHz
using a straightforward C++ implementation (no parallelism, no use of
GPUs).

\paragraph{$a^nb^n$ problem.} In this problem, the training sequence is
made of lines separated by newlines. The first line is a block of
$n_1$ symbols \texttt{a}; the second line is a block of $n_1$ symbols
\texttt{b}; the third line contains $n_2$ \texttt{a}, the fourth line
contains $n_2$ \texttt{b}, etc\@. See Example~\ref{ex:anbn}.

In this experiment, the block lengths $n$ were taken
at random in $\intint{1024;2048}$ to build the training and validation
sequences.

We used training and validation sequences made of only ten
$\texttt{a}^n\texttt{b}^n$
blocks.

\begin{figure}
\begin{centering}
\includegraphics{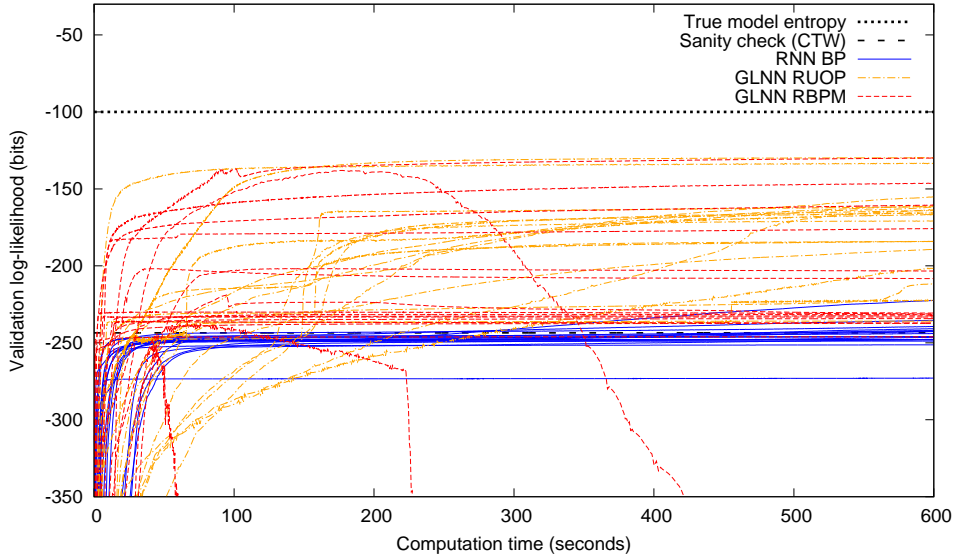}
\caption{Validation log-likelihood on the $a^nb^n$ example.}
\label{fig:anbn}
\end{centering}
\end{figure}

RNNs and GLNNs with sizes ranging from 4 to 64, as described above,
were run for 10 minutes each. For each independent run, a new random
training sequence and validation sequence was generated. The results are
reported in Figure~\ref{fig:anbn} and Table~\ref{fig:res}.

The log-likelihood of a validation sequence under the true model is 10
bits for each block of \texttt{a} (choosing an integer $n$ between $1024$
and $2048$), after which the length next block of \texttt{b} is known and
comes for free. Thus the reference log-likelihood of the whole validation
sequence (which contains 10 blocks of each) is 100 bits. However, from
only 10 training samples as used here, the exact distribution of the length $n$ cannot 
reasonably be inferred; a reasonable inference would be, for instance, a
geometric law with mean somewhere in this interval. The geometric
law with mean $\frac{1024+2048}{2}=1536$ has an entropy of about
12 bits instead of 10.

Thus, at best, one can expect a reasonable model to attain an entropy of
about 120 bits on the 10-instance-long validation set. On the other hand, a
model which would not catch the equality of the sizes of consecutive
\texttt{a} and \texttt{b} blocks would require twice as much entropy,
i.e., about 240 bits for the validation set. Indeed, the sanity check
(CTW) has a log-likelihood of -243.5 bits.

The best GLNN log-likelihood value obtained is -129.7 bits, while the best RNN
log-likelihood value is -222.4 bits.

Surprisingly, the best GLNN value was obtained with a network of size 4;
a size-23
network came close second at -129.98 bits.

Not all GLNN runs find the optimum: there is a cluster of runs around -230
bits, presumably corresponding to the model with independent lengths for
\texttt{a} and \texttt{b} blocks, and one run (with 64 units) provided
aberrant validation log-likelihood after some point because of
overfitting.

Visual inspection of the output of the best trained GLNN runs, used as
generative models, shows that consecutive blocks of \texttt{a} and
\texttt{b} indeed have the same or very close lengths, with sometimes an
error of $\pm 1$ on the length. This imperfection would likely disappear
with more than ten training sequences.

The kind of internal representation used by the GLNN to reach this
result is unclear, especially given the small network size: does it
build a kind of base-2 counter, does it take advantage of the analog
nature of the units' activities, or something in between?

\paragraph{Influence of the various choices.}
The difference in performance between GLNNs and RNNs above results from
various factors: choices in model design (leakiness and gatedness)
and in the training method (backpropagation or a Riemannian gradient). We
now try to isolate these factors, by testing various combinations of
models (RNNs, GNNs and GLNNs) and training methods.

In particular, it is possible to apply invariant training methods to
RNNs. The recurrent BPM and recurrent UOP metric
are well-defined for RNNs. However, contrary to GNNs and GLNNs, the
parameters corresponding to different symbols in the alphabet are not
mutually orthogonal, and thus, using them directly would result in a
complexity quadratic in the alphabet size, which we deem unacceptable.
Therefore, we used the quasi-diagonal reductions of these metrics, as
defined in \cite{gradnn}. This still provides training methods that are
invariant under reparametrization of the activity of each unit.

Each model and training method was tested as described above, spanning
various values of the hyperparameters (network size and connectivity).
For each method we report the best performance found over the
hyperparameters.

\begin{table}
\begin{center}
\begin{tabular}{lrrr}
\multicolumn{1}{c}{\multirow{2}{*}{Method}}
& \multicolumn{3}{c}{Cumulative regret (bits)}
\\
\cline{2-4}
& Alphabet & Music & $a^nb^n$
\\
\hline
\emph{Sanity checks:}
\\\texttt{bzip2}
& 27206.1 & 6035.7 & 244.0
\\Context tree weighting
& 18272.1 & 3611.0 & 143.5
\\
Hidden Markov model & 10212.0 & 4818.1 & 125.7
\\\hline
\emph{Non-invariant methods:}
\\
RNN with DH and FB & 7616.8 & 2898.9 & 122.4
\\
RNN with DH and RMS & 6721.5 & 2295.0 & 129.9
\\
GNN with DH and FB & 4059.0 & 2688.7 & 132.0
\\
GLNN with DH and FB & 5883.9 & 3616.4 & 74.4
\\
LSTM with plain gradient & 20073.1 & 8796.1 & 139.8
\\
LSTM with DH and FB & 11843.1 & 5590.9 & 101.7
\\ \emph{Invariant methods:}
\\
RNN with QDH and QDRUOP &  3372.8 & 1349.5 & 89.5
\\
RNN with QDH and QDRBPM & 3623.9 & 1224.3 & 106.0
\\
GNN with QDH and RUOP & 1759.0 & 1148.3 & 71.6
\\
GNN with QDH and RBPM & 2166.0 & 1596.0 & 115.4
\\
GLNN with QDH and RUOP & 1011.9 & 1055.9 & 29.7
\\
GLNN with QDH and RBPM & 644.2 & 1055.9 & 30.0
\\\hline
\end{tabular}
\end{center}
\caption{Cumulative regret (bits) of the learned model with respect to
the true generative model, over the validation data sequence. Best value
found over the hyperparameters in the allocated CPU time budget is
reported. Models are: RNN, GNN, GLNN, LSTM,
and HMM. Training methods for the writing parameters: inverse diagonal
Hessian (DH), or quasi-diagonal inverse Hessian (QDH). Training methods
for the transition parameters: frequency-adjusted backpropagation (FB),
root mean square gradient rescaling (RMS),
recurrent backpropagated metric (RBPM), recurrent unitwise outer product
metric (RUOP),
and the quasi-diagonal reduction of the latter (QDRBPM and QDRUOP).
}
\label{fig:res}
\end{table}

The performance reported is the \emph{cumulative regret} with respect to the
true generating model, a standard measure used in sequential prediction
contexts. It is defined as the difference between the log-likelihood of
the validation data sequence under the true model used to generate the
data, and the log-likelihood of the validation data sequence under the
trained model.

We also included three sanity checks for reference. Two are text compressors
known for their performance (CTW as mentioned above, and the file
compressor \texttt{bzip2}), for which, to incorporate the effect of
training, we report the number of bits used to compress the concatenation
of the training and validation sequences minus the number of bits used to
compress the training sequence alone.

The third sanity check is a hidden Markov model (HMM), trained using a variety
of network sizes as for the neural networks.\footnote{Details as follows.
Training is done by the expectation-maximization algorithm. The network
is obtained, as for the neural networks, by taking an oriented random graph with a
given number of edges per node (including loops); this number of edges
per node is set to the
alphabet size, because this gives an algorithmic complexity similar to that of
the neural networks. Initialization of the transition probabilities is by
a Dirichlet$(1/2,\ldots,1/2)$ (i.e., Jeffreys) prior on the edges from a node.
Initialization of the production probabilities is done by multiplying the
actual frequency of each symbol in the sequence to be modelled, by a
random uniform$([0;1])$ number.} The comparison with HMMs is especially
interesting, since these are a classical tool for modelling sequential
data.

The ``classical'' training method is as described above for RNNs:
diagonal inverse Hessian for the writing parameters $w$, and
backpropagation for the transition parameters; for the latter, parameters
like $\rho_{iy}$ (for RNNs) or $\tau_{ijy}$ (for GNNs and GLNNs) related
to a given symbol $y$ have a learning rate divided by the frequency of
$y$ in the training sequence (``frequency-adjusted'' backpropagation, which compensates for the number of terms
making up the corresponding gradient, and, for RNNs, is equivalent to
scaling the input signals). Pure backpropagation
was tested but is simply too slow.

The results are collected in Table~\ref{fig:res}.

From this table it is clear that an invariant method is the first step to
improve performance: RNNs trained with an invariant method beat GNNs and
GLNNs trained with a non-invariant method.

Still, the leaky aspect of GLNNs seems to be necessary to
bring the best performance in problems with very long dependencies (the
alphabet with insertions and the $a^nb^n$ example). On the other hand, on
the problem where dependencies are most local (synthetic music), all
network models achieve quite comparable results if trained with an
invariant method.

\section*{Conclusions and perspectives}

The viability of GLNNs with Riemannian training to capture complex
algorithmic dependencies in symbolic data sequences has been established.
Metrics inspired by a Riemannian geometric viewpoint, allow us to write invariant algorithms at an
algorithmic cost comparable to backpropagation through time for sparsely
connected networks.

These metrics bring down the necessary number of gradient steps to a few
hundreds in the various examples studied. This approach seems to work
with small training samples. Better than state-of-the-art
performance has been obtained on difficult synthetic problems.

In the experiments, the importance of invariance seems to supercede that
of model choice: in our tests, any model with an invariant training
algorithm did better than any model with a non-invariant one.

More experiments are needed to investigate the isolated effect of each
feature of this training procedure (memory effect in the definition of
GLNNs, gatedness, and the choice of metric).
Other issues in need of investigation
are the influence of parameter initialization (especially if some expert
knowledge on the time scale of dependencies in the data is available) and
a better, invariant dampening procedure. It would also be interesting to acquire
insight into the dynamical behavior of GLNNs (ergodicity, multiple
equilibrium regimes, etc.) and how it is affected by training. Furthermore, the Riemannian approach can in
principle be extended to more complex architectures: testing
Riemannian methods for LSTMs seems promising.

Finally, 
scalable Riemannian training algorithms should be developed for a fully
online ``lifelong learning''
setting where there is a single training sequence which grows with time and where
it is not possible to fully store the past states and signal, so that
backpropagation through time is excluded.

\appendix

\section{Parameter initialization, the 
linearized regime, and integrating effects in GLNNs}
\label{sec:init}

Let us examine the dynamics of a GLNN, and in particular the
linearized regime (the regime in which the connection weights are small).
This will provide some insight into the time-integrating effect of the
model, and also suggest relevant initializations of the parameter values
before launching the gradient ascent, as presented in the algorithm
above.

In the GLNN evolution equation
$V^{t+1}_j= V^t_j+\sum_i \tau_{ijx_t}a^t_i$
let us isolate the contributions of $i=j$ and of the always-activated
unit $i=0$. Substituting $a^t_j=\actf(V^t_j)$ and $a^t_0\equiv 1$ we get
\begin{equation}
V^{t+1}_j= V^t_j+\tau_{jjx_t}\actf(V^t_j)+\tau_{0jx_t}+\sum_{i\neq
0,\,i\neq j} \tau_{ijx_t}a^t_i
\end{equation}

Since $\actf(V^t_j)$ is an increasing function of $V^t_j$, the
contribution $i=j$ provides a feedback loop: if
$\tau_{jjx}$ is negative for all $x$, then the feedback will be negative,
whereas positive $\tau_{jjx}$ would result in perpetual increase of
$V^t_j$ if the other contributions are ignored. Meanwhile,
$\tau_{0jx_t}$ provides the reaction of unit $j$ to the signal $x_t$.

For instance, if we set
$\tau_{jjx}=-\alpha$ for all $x$ with $\alpha>0$, $\tau_{0jx}=\beta$ for all $x$, 
and all other
weights $\tau_{ijx}$ to $0$, the dynamics is
\begin{equation}
V^{t+1}_j= V^t_j-\alpha \actf(V^t_j) +\beta
\end{equation}
which has a fixed point at $V^t_j= \bar V\deq \actf^{-1}(\beta/\alpha)$, i.e.,
$a^t_j= \beta/\alpha$ (assuming $\beta/\alpha$ lies in the range of
the activation function $\actf$). The linearized
dynamics around this fixed point ($V^t_j$ close to $\bar V$) is
\begin{equation}
V^{t+1}_j-\bar V \approx (1-\alpha \actf'(\bar V))\left(V^t_j-\bar V\right)
\end{equation}
so that if $\abs{1-\alpha \actf'(\bar V)}<1$ this fixed point is
attractive.

A more interesting choice is to let
\begin{equation}
\tau_{jjx}=-\alpha,\qquad
\tau_{0jx}=\beta+\eps\rho_{x}
\end{equation}
with small $\epsilon$, where $\rho_x$ is chosen to that the average of
$\rho$ over the data $x_t$ is $0$. Then, the value of $V^t$ as a function
of $t$ and the data can be found
by induction using the linearized dynamics:
\begin{equation}
V^t_j\approx \bar V+\eps\sum_{t'<t} (1-\mu)^{t-t'}\rho_{x_{t'}}
\end{equation}
where
\begin{equation}
\mu\deq\alpha \actf'(\bar V)
\end{equation}
namely, the activation level $V^t_j$ is a linear combination of the past
values of the signal $x_t$, with weights exponentially decreasing with
time at rate $(1-\mu)$.

This provides insights into reasonable values of the parameter leading to
interesting internal dynamics, to be used at the start of the learning
procedure. Indeed, negative values of $\alpha$ would lead to
unstability, whereas positive values of $\alpha$ presumably stabilize the
network. However, values of $\alpha$ above $1/\sup \actf'$ ($=1$ for $\tanh$
activation) will provide too much
feedback, resulting in
non-monotonous $V^{t+1}$ as a function of $V^t$ and an oscillating
behavior. Indeed we have found that setting $\alpha=1/(2\sup \actf')$,
i.e.,
\begin{equation}
\tau_{jjx}=-\alpha=-1/2
\end{equation}
(for tanh) for all $j$ and $x$ at startup, yields a good behavior of the network.

With $\tau_{jjx}$ and $\tau_{0jx}$ as above, the value of $(1-\mu)$ controls
the effective time window of the integrating effect: data much older than
$t-t'\gg \frac{1}{\mu}$ has little weight. Thus $\frac{1}{\mu}$ presumably gives
the order of magnitude of the distances $t-t'$ for which the model can
reasonably be expected to capture correlations in the data (at least at startup,
since $\mu$ will change during learning).

The value of $\mu$ can be directly controlled through $\beta$ via
$\mu=\alpha \actf'(\bar V)=\alpha \actf'(\actf^{-1}(\beta/\alpha))$:
for the $\tanh$ activation function, this yields
\begin{equation}
\beta=-\sqrt{\alpha(\alpha-\mu)}
\end{equation}
which is used to set $\beta$ from an arbitrary choice for $\mu$.
We have found
that using different values of $\mu$ for different units yields good
results. We have used 
\begin{equation}
\mu_j=1/(j+1)
\end{equation}
for unit
number $j$ (starting at $j=1$); this yields a characteristic time 
of order $j$ and seems to perform well.

Finally, the ``reading rates'' $\rho_x$ are taken at random independently
for each unit $j$ in the following way. The value of $\eps$ must be small
enough to ensure that $V^t_j$ stays close to $V_j$ (otherwise the linear
regime assumption is unjustified), namely, that the sum $\eps \sum_{t'<t}
(1-\mu)^{t-t'}\rho_{x_{t'}}$ stays small. If each $\rho$ is roughly of
size $1$, the sum is $\eps/\mu$ so taking $\eps$ somewhat smaller
than $\mu$ is reasonable. We have used
\begin{equation}
\eps=\frac{\mu}{4}
\end{equation}
which apparently yields good performance. Finally, $\rho_x$ is taken at random
uniformly in $[0;1]$ for each symbol $x$ (independently for each unit
$j$), and then shifted by a constant so that the average of $\rho_{x_t}$
over the training data $x_t$ is $0$ (namely, the constant $\sum \nu_x
\rho_x$ is removed from each $\rho_x$ where $\nu_x$ is the frequency of
symbol $x$ in the training data)\footnote{The choice to use a uniform random
variable in $[0;1]$ rather than, e.g., Gaussian random variables, is
justified by the feedback mechanism. Indeed since the activation function
$\actf$ ranges in $[0;1]$, the feedback term $-\alpha s(V^t_j)$ is
bounded. If an unbounded signal $\rho_{x_t}$ can occur at each step, it may take a long
time to stabilize. Empirically, using Gaussian rather than bounded random
variables seems to decrease performance, confirming this viewpoint.}.

The other transition weights $\tau_{ijx}$, with $i\neq 0$, $i\neq j$, were
set to $0$ at startup.

The explicit initialization values described here are specific to the
tanh activation function; however, the reasoning extends to
any activation function.

\section{Derivative of the log-likelihood: Backpropagation
through time for GLNNs}
\label{sec:backprop}

Let $(x_t)_{t=0,\ldots,T-1}$ be an observed sequence of $T$ symbols in
the alphabet $\A$. Here we compute the derivatives of the log-probability
that a GLNN prints $(x_t)$ with respect to the GLNN parameters, via the
standard backpropagation through time technique.

Given a training sequence $x=(x_t)$, let $\Pr(x)$
be the probability that the model prints $(x_0,\ldots,x_{T-1})$. Here,
for simplicity we assume that all symbols in the sequence have to be
predicted (i.e., $\chi_t\equiv 1$). The algorithm in Section~\ref{sec:algo} gives the
formulas for the general case.

\begin{prop}[ (log-likelihood derivative for GLNNs)]
\label{prop:glnnder}
The derivative of the log-probability of a sequence
$x=(x_t)_{t=0,\ldots,T-1}$ with respect to the parameters of a gated
leaky
neural network is given as follows.

Setting
\begin{equation}
B^t_j\deq \frac{\partial \log \Pr(x)}{\partial V^t_j}
\end{equation}
we have the backpropagation relation
\begin{equation}
B^t_i=B^{t+1}_i+\actf'(V^t_i)
\left(
w_{ix_t}-\sum_y \pi_t(y)w_{iy}+\sum_j \tau_{i j
x_t} B^{t+1}_j
\right)
\end{equation}
(initialized with $B^T_j\deq 0$). In particular $B^0_j$ gives the
derivative with respect to the initial values $V^0_j$ at time $0$.

The derivatives with respect to the writing weights are
\begin{equation}
\label{eq:gradw}
\frac{\partial \log \Pr(x)}{\partial w_{iy}}=
\sum_{t} a_i^t \left(\1_{x_t=y}-\pi_t(y)\right)
\end{equation}
and the derivatives with respect to the transition weights are
\begin{equation}
\frac{\partial \log \Pr(x)}{\partial \tau_{ijy}}=
\sum_t \1_{x_t=y}\,
a^t_i B^{t+1}_j
\end{equation}

\end{prop}

These relations include the always-activated unit $i=0$, $a_i\equiv 1$.

The meaning of the partial
derivative with respect to $V^t_j$ is the following: if, in the equation
$V^{t+1}_j= V^t_j+\sum_j \tau_{ijx_t}a^t_i$ defining GLNNs, we
artificially introduce a term $\eps\ll 1$ at unit $j$ at time $t$,
namely, $V^{t+1}_j= V^t_j+\sum_j \tau_{ijx_t}a^t_i+\eps$ for a single
unit at a single time,
and let the network evolve normally except for this change, then the
value of $\log P_T$ changes by $\eps B^t_j +O(\eps^2)$.

\begin{proof}
Given a training sequence $(x_t)_{t=0,\ldots,T-1}$ of length $T$,
let $P_0\deq 1$ and
\begin{equation}
P_{t+1} \deq \pi_t(x_t)P_t
\end{equation}
so that $P_T$ is the probability of printing $(x_0,\ldots,x_{T-1})$.

By definition of $\pi_t$ we have
\begin{equation}
\log P_{t+1}=\log P_i+\sum_i a^t_i w_{ix_t}-\log\left(
\sum_y \e^{\sum_i 
a^t_i w_{iy}}
\right)
\end{equation}

Let us compute the infinitesimal variations of these quantities under an
infinitesimal variation $\d
w$, $\d\tau$ of the parameters. Ultimately we are interested in the variation of
$\log P_T$, to perform gradient ascent on the parameters.

By a first-order Taylor expansion, the variation of $\log P_{t+1}$ satisfies
\begin{equation}
\begin{split}
\d\log P_{t+1} ={} &
\d\log P_t +\sum_i a^t_i \d w_{i x_t}+\sum_i w_{ix_t}\d a^t_i
\\& -\sum_y \pi_t(y) \left(
\sum_i a^t_i \d w_{i y}+\sum_i w_{iy}\d a^t_i
\right)
\end{split}
\end{equation}
and rearranging and substituting
\begin{equation}
\d a_i^t=\actf'(V^t_i)\d V^t_i
\end{equation}
where $\actf'$ is the derivative of the activation function, this yields
\begin{equation}
\label{eq:Ptind}
\begin{split}
\d\log P_{t+1} ={} &
\d\log P_t +\sum_i a^t_i \left(
\d w_{i x_t}-\sum_y \pi_t(y) \d w_{iy}\right)
\\&
+\sum_i \left(w_{ix_t}
-\sum_y \pi_t(y)w_{iy}\right) 
\actf'(V^t_i)\d V^t_i
\end{split}
\end{equation}

Consequently, the variation $\d\log P_t$ of $\log P_t$ can be expressed
in terms of the variation of $\log P_{t-1}$, the variations of the
parameters $w$ and $\tau$, and the variations of the values $V^{t-1}_j$
at time $t-1$.

Let us assume, by backward induction, that we can write the differential
of $\log P_T$ with respect to the parameters, as
\begin{equation}
\label{eq:tderind}
\d\log P_T \eqd \d\log P_t 
+
\sum_i B^t_i \d V^t_i
+\sum_{i,y} C^t_{iy} \d w_{iy}
+\sum_{i,j,y} D^t_{ijy}\d\tau_{ijy}
\end{equation}
For $t=T$ this is satisfied with $B^T=C^T=D^T=0$.

Thus $B^t_i$ represents the
backpropagated value at unit $i$ at time $t$, and $C$ and $D$ will
cumulatively compute the gradient of $\log P_T$ with respect to the
parameters $w$ and $\tau$, namely:
\begin{equation}
\frac{\partial \log \Pr(x)}{\partial w_{iy}}=C^0_{iy}
\end{equation}
and
\begin{equation}
\frac{\partial \log \Pr(x)}{\partial \tau_{ijy}}=D^0_{ijy}
\end{equation}
and moreover $B^0_j$ will contain the derivatives with respect to the initial
levels $V^0_j$.

Using the evolution equation $V^{t+1}_j=V^t_j+\sum_i \tau_{ijx_t} a^t_i$
we find
\begin{equation}
\d V^{t+1}_j=\d V^t_j+\sum_i \d\tau_{ijx_t} a^t_i+\sum_i
\tau_{ijx_t}\actf'(V^t_i)\d V^t_i
\end{equation}

Using these relations to go from time $t+1$ to time $t$
in~\eqref{eq:tderind}, namely, expressing $\d\log P_{t+1}$ in terms of
$\d\log P_{t}$ and
expanding $V^{t+1}$ in terms of $V^t$, we find
\begin{equation}
C^t_{iy}=C^{t+1}_{iy}+\1_{x_t=y}\,a^t_i-\pi_t(y)a^t_i
\end{equation}
\begin{equation}
D^t_{ijy}=D^{t+1}_{ijy}+\1_{x_t=y}\,
a^t_i B^{t+1}_j
\end{equation}
and
\begin{equation}
B^t_i=B^{t+1}_i+\actf'(V^t_i)
\left(
w_{ix_t}-\sum_y \pi_t(y)w_{iy}+\sum_j \tau_{i j
x_t} B^{t+1}_j
\right)
\end{equation}
from which the expressions for $C^0_{iy}$ and $D^0_{ijy}$ follow.
\end{proof}

\section{Fisher metric for the output distribution $\pi_t$}
\label{sec:fishout}

Let us compute the Fisher norm of the variation $\d\pi$ of $\pi$ resulting from
a change $\d E_y^t$ in the values of $E_y^t=\sum_i a^t_i w_{iy}$. (Such a change in $E$ can result
from a change in the writing weights $w$ or the activities $a$; this
will be used to compute the metric on the writing weights and the
transition weights, respectively.) The effect of a change $\d E^t$ on
$\log \pi_t$
is
\begin{equation}
\d\log \pi_t(y)=\sum_{y'} \frac{\partial \log \pi_t(y)}{\partial E^t_{y'}}
\d E^t_{y'}
\end{equation}
and the norm of this $\d\pi_t$ in Fisher metric is
\begin{align}
\natnorm{\d\pi_t}&=\E_{y\sim \pi_t} (\d\log \pi_t(y))^2
\\&=
\E_{y\sim\pi_t} \left[\sum_{y',y''} \frac{\partial \log \pi_t(y)}{\partial
E_{y'}}\frac{\partial \log \pi_t(y)}{\partial E_{y''}}\d E^t_{y'}\d
E^t_{y''}\right]
\end{align}

By a standard formula for
exponential families of probability distributions we find:
\begin{equation}
\frac{\partial \log \pi_t(y)}{\partial E_{y'}}=\1_{y=y'}-\pi_t(y')
\end{equation}
so that
\begin{align}
\E_{y\sim\pi_t} \left[\frac{\partial \log \pi_t(y)}{\partial
E_{y'}}\frac{\partial \log \pi_t(y)}{\partial E_{y''}}\right]
&=
\E_{y\sim\pi_t}\left[(\1_{y=y'}-\pi_t(y'))(\1_{y=y''}-\pi_t(y''))\right]
\\&=\pi_t(y')(\1_{y'=y''}-\pi_t(y''))
\end{align}
(this is also\footnote{because for exponential families, the Hessian of $\log
\pi(y)$ does not depend on $y$} the Hessian of $-\log \pi_t(y)$ with
respect to the values $E^t$).
Consequently, the Fisher metric for $\pi_t$, expressed in terms of the
variations $\d E_y^t$, is
\begin{equation}
\natnorm{\d\pi_t}^2=\sum_y \pi_t(y)(\d E_y^t)^2-\sum_{y,y'}\pi_t(y)\pi_t(y')\d E_y^t\d
E_{y'}^t
\end{equation}

%
%


\bibliographystyle{alpha}
\bibliography{pcnn}

\end{document}